\documentclass[10pt,twocolumn,letterpaper]{article}

\usepackage{cvpr}              %

\newif\ifarxiv
\arxivtrue %

\usepackage[dvipsnames]{xcolor}

\usepackage[accsupp]{axessibility} %
\usepackage{graphicx}
\usepackage{tikz}
\usepackage{comment}
\usepackage{amsmath,amssymb,amsfonts}
\usepackage{color}
\usepackage{booktabs}
\usepackage{tabularx}
\usepackage{arydshln}

\usepackage{siunitx}
\usepackage{etoolbox}
\robustify\bfseries

\usepackage{makecell}
\usepackage{adjustbox} %
\usepackage{array}
\usepackage{multirow}
\usepackage{rotating}

\usepackage{caption}
\usepackage{subcaption}

\def\pol{p} %

\def\pol{p}

\def\ray{\mathrm{r}}

\newcommand{\bnum}[1]{\bfseries #1}

\definecolor{light-gray}{gray}{0.6}

\newcommand{\argmin}{\mathop{\rm arg~min}\limits}

\usepackage[absolute]{textpos}

\definecolor{cvprblue}{rgb}{0.21,0.49,0.74}
\usepackage[pagebackref,breaklinks,colorlinks,citecolor=cvprblue]{hyperref}
\usepackage{orcidlink}

\usepackage{url}
\usepackage[capitalize]{cleveref}
\crefname{section}{Sec.}{Secs.}
\Crefname{section}{Section}{Sections}
\Crefname{table}{Table}{Tables}
\crefname{table}{Tab.}{Tabs.}

\def\paperID{5} %
\def\confName{CVPR}
\def\confYear{2024}

\ifarxiv
\usepackage{orcidlink}
\usepackage{balance}
\usepackage[absolute]{textpos}
\hypersetup{
  pdftitle={3D Human Scan With A Moving Event Camera},
  pdfsubject={Computer Vision, Robotics},
  pdfauthor={Kai Kohyama, Shintaro Shiba, Yoshimitsu Aoki},
  pdfkeywords={Event camera, Human Pose Estimation, Human Mesh Recovery}
}
\makeatletter
\long\def\@IEEEtitleabstractindextextbox#1{\parbox{0.922\textwidth}{#1}}
\makeatother
\fi

\title{3D Human Scan With A Moving Event Camera}

\ifarxiv
\author{Kai Kohyama\orcidlink{0009-0005-4090-8819}\\
{\tt\small kaikohyama@keio.jp}
\and
Shintaro Shiba\orcidlink{0000-0001-6053-2285}\\
{\tt\small sshiba@keio.jp}\\
Keio University
\and
Yoshimitsu Aoki\orcidlink{0000-0001-7361-0027}\\
{\tt\small aoki@elec.keio.ac.jp}\\
}
\else
\author{Kai Kohyama\\
{\tt\small kaikohyama@keio.jp}
\and
Shintaro Shiba\\
{\tt\small sshiba@keio.jp}\\
Keio University
\and
Yoshimitsu Aoki\\
{\tt\small aoki@elec.keio.ac.jp}\\
}
\fi 

\begin{document}

\ifarxiv
\definecolor{somegray}{gray}{0.6}
\newcommand{\darkgrayed}[1]{\textcolor{somegray}{#1}}
\begin{textblock}{11}(2.5, 0.4)
\begin{center}
\darkgrayed{This paper has been accepted for publication at the\\
IEEE Conference on Computer Vision and Pattern Recognition (CVPR)\\
Workshop On Computer Vision For Mixed Reality (CV4MR), Seattle, 2024.
\copyright IEEE}
\end{center}
\end{textblock}
\fi 

\maketitle
\begin{abstract}

Capturing the 3D human body is one of the important tasks in computer vision with a wide range of applications such as virtual reality and sports analysis.
However, conventional frame cameras are limited by their temporal resolution and dynamic range, which imposes constraints in real-world application setups.
Event cameras have the advantages of high temporal resolution and high dynamic range (HDR),
but the development of event-based methods is necessary to handle data with different characteristics.
This paper proposes a novel event-based method for 3D pose estimation and human mesh recovery.
Prior work on event-based human mesh recovery require frames (images) as well as event data.
The proposed method solely relies on events;
it carves 3D voxels by moving the event camera around a stationary body,
reconstructs the human pose and mesh by attenuated rays,
and fit statistical body models, preserving high-frequency details.
The experimental results show that the proposed method outperforms conventional frame-based methods
in the estimation accuracy of both pose and body mesh.
We also demonstrate results in challenging situations
where other frame-based methods suffer from motion blur.
This is the first-of-its-kind to demonstrate event-only human mesh recovery,
and we hope that it is the first step toward achieving
robust and accurate 3D human body scanning from vision sensors.

\end{abstract}

\section{Introduction}
\label{sec:intro}

\def\figWidth{1.0\linewidth}
\begin{figure}[t]
    \begin{center}{
        {\includegraphics[trim={0.5cm 3.5cm 1.5cm 5cm},clip,width=\figWidth]{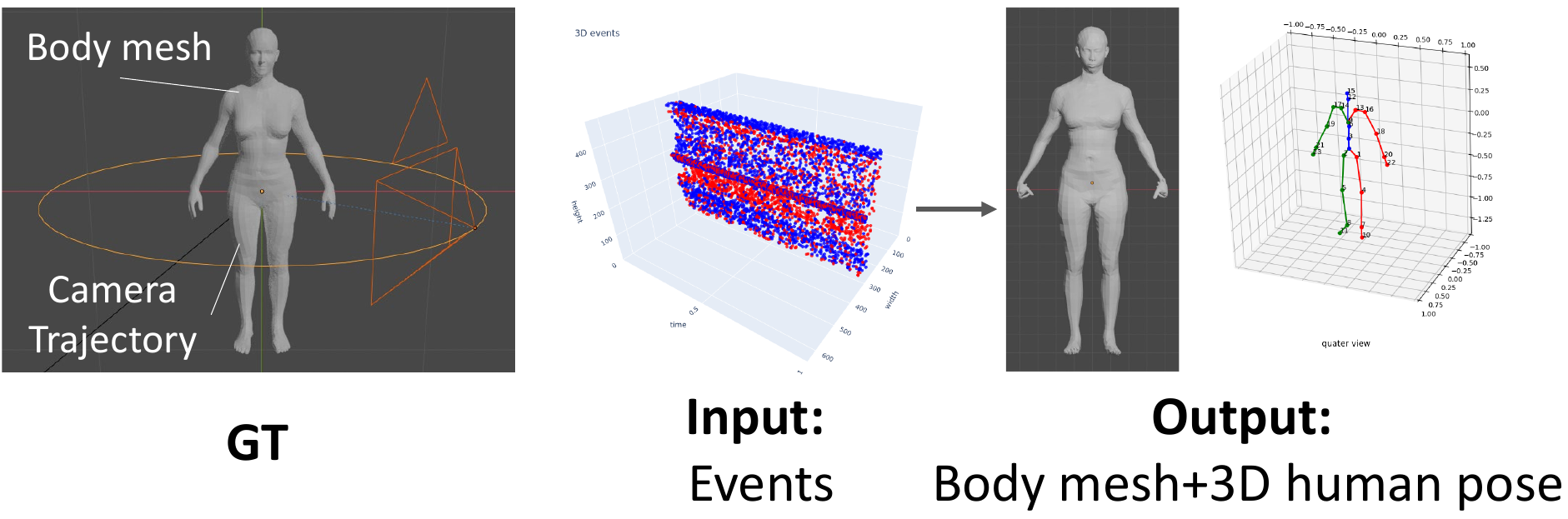}}
        \caption{Summary of the proposed method.
        Our method reconstructs the human body mesh and estimates the pose only from an event camera that moves around the body.
        }
        \label{fig:overview}
        }
    \end{center}
\end{figure}

\def\figWidth{0.95\linewidth}
\begin{figure*}[t!]
    \begin{center}
        {\includegraphics[trim={1.2cm 14cm 0.5cm 4.2cm},clip,width=\figWidth]{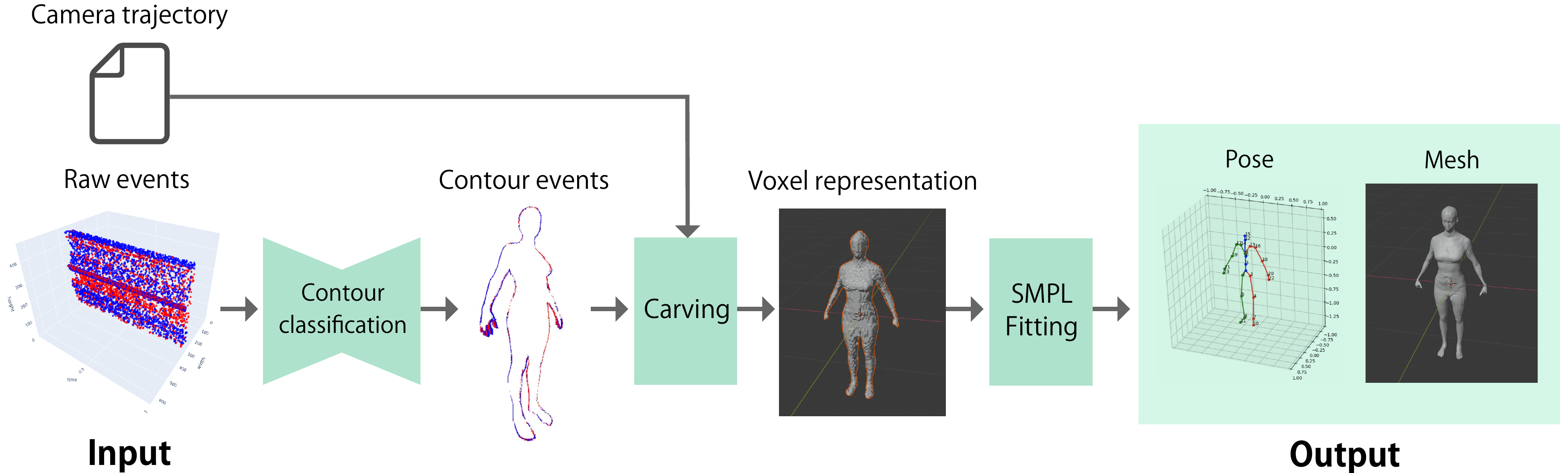}}
        \caption{Overview of the proposed method.
        }
        \label{fig:event_based_method}
    \end{center}
\end{figure*}

Estimating human pose from cameras is one of the key computer vision challenges with a wide range of applications
such as virtual reality (VR), sports analysis, and abnormal behavior detection.
Recently, many pose estimation methods using deep neural networks (DNNs) have been proposed and developed
for images from conventional frame cameras.
However, such methods have limitations in the applicable scenes
 that inherits the constraints of frame cameras by nature:
the temporal resolution is insufficient for intense motion such as during sports (i.e., motion blur), 
and the shutter speed must be adjusted to obtain data in dark scenes (i.e., limited dynamic range).
To address these challenges, event cameras have recently gained attention from both research and industry.
Unlike conventional frame cameras where all pixels record data synchronously,
event cameras asynchronously respond only to brightness changes,
resulting in high temporal resolution ($\si{\micro\sec}$ order) and high dynamic range (HDR).
However, since event cameras produce different data from frames \cite{Gallego20pami,Lagorce17pami},
it is paramount to develop event-based methods for pose estimation and mesh recovery using event cameras.
Previous research on human body mesh reconstruction using event cameras \cite{eventCap2020CVPR, zou2021eventhpe}
require frame images as complementary information,
and could not achieve reconstruction from events alone.
This is because event cameras do not generate data for stationary parts of the body
if the camera is static (i.e., no event data observed).
Frame cameras are better for capturing such static scene information,
however, using frame images is problematic
since it imposes the limitations of frames (e.g., dynamic range, motion blur).
We summarize existing frame-based and event-based methods in \cref{tab:intro_comp_methods}.
\begin{table}[t]
\centering
\caption{Comparison of (some) existing methods for frame-based and event-based human mesh reconstruction. Scene and Camera can be either dynamic (``D'') or static (``S'').
Existing event-based methods require images (``I'') as well as events (``E''), resulting in mitigating motion blur to some extent (denoted with $\dagger$).}
\label{tab:intro_comp_methods}
\setlength{\tabcolsep}{3pt}
\begin{tabular}{ccccc}
\hline
& Scene & Camera & Data  & Motion blur
\\ \hline
PyMAF \cite{pymaf2021} & D & S  & I & Yes \\
PyMAF-X \cite{pymafx2023} & D & S  &  I & Yes \\
EasyMoCap \cite{pymaf2021} & D & S  & I & Yes \\
\hline
EventCap \cite{eventCap2020CVPR} & D & S  & E $+$ I & No$\dagger$ \\
EventHPE \cite{zou2021eventhpe} & D & S  &  E $+$ I & No$\dagger$ \\
Ours & S & D & \textbf{E} & \textbf{No} \\ 
\hline
\end{tabular}
\end{table}

In this work,
we propose a 3D human body scanning method capable of estimating the 3D voxel representation, mesh, joints, and body model parameters of a stationary human body using only event data (\cref{fig:overview}).
Our proposed method enables data acquisition from static bodies by moving the camera itself.
Furthermore, we propose a ray attenuation to better preserve high-frequency detail information,
by extending the existing event-based voxel carving method \cite{Wang2022EvAC3D}.
Finally, by fitting the statistical human body models, such as SMPL \cite{SMPL:2015} and SKEL \cite{keller2023skel},
we demonstrate accurate estimation of the human body mesh, body parameters, and joint positions from the voxels.

To summarize, our contributions are as follows:
\begin{itemize}
    \item The first-of-its-kind method to estimate human pose and body mesh from only events.
    \item A ray attenuation for preserving details better, compared with existing event-based carving.
    \item A thorough benchmarking with frame-based methods and ablation studies on different illuminations, camera speeds, hyper-parameters, and statistic body models.
\end{itemize}
The experimental results indicate that the proposed method achieves accurate human body mesh reconstruction and 3D pose estimation from events alone,
without the need for additional frames.
Leveraging the high temporal resolution of event cameras,
our method achieves precise carving surpassing frame-based pose estimation methods.

\section{Related Work}
\label{sec:related}

\subsection{Frame-based Human Pose and Mesh Estimation}

3D human pose estimation from frame-based cameras, which involves inferring the positions of human joints,
is one of the popular tasks in computer vision.
Most of the recent work utilize deep learning in various manners,
such as direct regression of 3D joints from images \cite{li2015maximum, park20163d, sun2017compositional},
triangulating two-dimensional pose estimation results into the 3D space \cite{chen20173d, li2019generating, wang20193d},
and estimating joint positions using heatmaps after converting the human body into a 3D representation \cite{pavlakos2017coarse, moon2018v2v, trumble2018deep}.
These approaches also consist of a wide range of problem settings,
from using a single camera \cite{park20163d, chen20173d, pavlakos2017coarse}
to employing multiple cameras \cite{trumble2018deep, dong2019fast,easymocap,Shuai22siggraph},
and utilizing depth sensors as additional information \cite{moon2018v2v, yu2018doublefusion}.

While traditional pose estimation focuses solely on the joint positions,
recent work have addressed human body \emph{mesh} reconstruction \cite{hmrKanazawa17,Kocabas20cvpr,pymaf2021,pymafx2023}.
In the mesh reconstruction,
not only the pose parameters but also the ones related to the body shape are estimated simultaneously to obtain the body mesh corresponding to the input image.
Meshes are generated by parametric statistical human body models,
such as SMPL \cite{SMPL:2015},
SMPL-X that extends SMPL for hands and faces \cite{SMPL-X:2019},
SKEL that considers anatomical structures \cite{keller2023skel},
and CAPE that includes clothes \cite{Ma20cvpr}.

\subsection{Event-based Human Pose and Mesh Estimation}

Human pose and mesh estimation using event cameras are relatively recent research fields due to the novelty of event cameras \cite{Gallego20pami}.
Common approaches accumulate events over a certain time interval,
convert them into images,
and feed them into convolutional neural networks (CNNs) to obtain the joint heatmaps \cite{Calabrese19cvprw, scarpellini2021lifting,zou2021eventhpe}.
Also, Chen et al. \cite{chen2022efficient} utilize point-cloud neural networks to directly process events,
reducing memory consumption and computational complexity.

Event-based human mesh reconstruction methods have also been studied,
such as the approach by Xu et al. \cite{eventCap2020CVPR}
that achieves $1000$~fps human body capture in low-light conditions.
However, it requires inputs from both events and frames,
as well as prior scanning of the subject to realize accurate reconstruction.
Zou et al. \cite{zou2021eventhpe} propose an approach using optical flow from events to estimate both pose and body mesh,
yet rely on frames for initialization.
Event-based hand mesh reconstruction method \cite{jiang2024complementing} also uses the complementary frame data,
and to the best of the authors' knowledge,
no work has shown the mesh recovery solely from events.
Our proposed method utilizes only event data (see \cref{tab:intro_comp_methods})
to simultaneously estimate the 3D joint positions
and reconstruct parametric human body models,
achieving precise estimation.

\subsection{Visual Hull}

Visual Hull, or carving, is a technique to reconstruct three-dimensional shapes by finely divided voxels, which are cubes carved from various angles based on the contours of objects extracted from images \cite{visualhull}.
In carving, a smoother reconstruction is achieved by increasing viewpoints.
Increasing the number of voxel subdivisions allows for finer detail representation, however, it requires more rays for sufficient carving,
leading to larger data size for the three-dimensional representation.
Moreover, due to its principle, high-frequency detailed information is difficult to extract contours,
and concave surfaces cannot be reconstructed.
Event cameras, due to their higher temporal resolution compared to traditional frame cameras, enable smoother voxel carving with more continuous viewpoint changes \cite{Wang2022EvAC3D}.
It demonstrates reconstructing smooth voxels for 3D reconstruction of simple objects like cans using an event camera.

\section{Method}
\label{sec:method}

In this section, we propose a method to estimate the 3D voxel representation, mesh, and parameters of joints (pose) and body shape for a stationary human body from a moving event camera. %
\Cref{fig:event_based_method} shows the overview of the proposed method.
It consists of three steps:
($i$) classifying ``contour'' events among the raw events (\cref{sec:method:classification}),
($ii$) carving voxels based on the contour events (\cref{sec:method:carving}),
and ($iii$) fitting statistical body models on the carved voxels (\cref{sec:method:regression}).
Our proposed pipeline is model-agnostic, as shown in \cref{sec:results}, and 
achieves accurate estimation thanks to the high temporal resolution.
\def\figWidth{1.0\linewidth}
\begin{figure}[t]
    \begin{center}
        {\includegraphics[trim={2.3cm 4.5cm 6.2cm 3.5cm},clip, width=\figWidth]{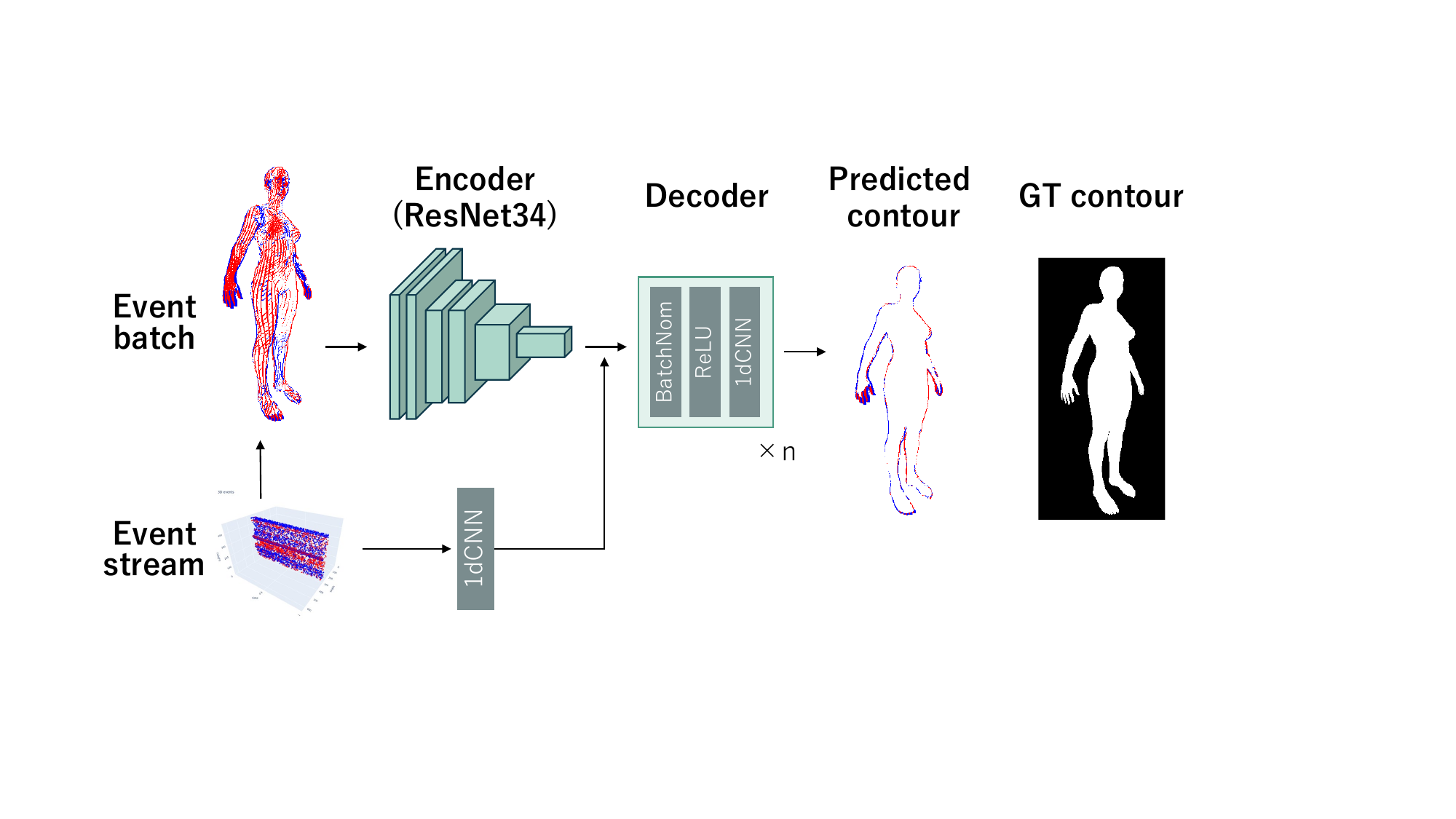}}
        \caption{Contour classification network.}
        \label{fig:classification_network}
    \end{center}
\end{figure}

\subsection{Event Cameras}
\label{sec:method:eventCam}

Event cameras (e.g., the Dynamic Vision Sensor (DVS) \cite{Lichtsteiner08ssc,Suh20iscas,Finateu20isscc})
are novel vision sensors that respond to intensity changes.
There have been emerging computer vision research for event cameras,
such as object detection \cite{Glover16iros,Mitrokhin18iros,Mueggler17bmvc},
ego-motion estimation \cite{Gallego18cvpr,Shiba22aisy},
optical flow \cite{Benosman12nn,Shiba22eccv,Shiba23spl},
SLAM \cite{Kim14bmvc,Kim16eccv,Guo24tro},
and many applications \cite{Lin20eccv,Muglikar21threedv,Angelopoulos20tvcg,Shiba23pami}.
They have independent pixels that operate continuously and generate ``events''
$e_k \doteq (x_k,y_k,t_k,\pol_k)$
whenever the logarithmic brightness at the pixel increases or decreases by a predefined sensitivity $C$:
\begin{equation}
\label{eq:event}
    L\left(x_k, y_k, t_k\right)-L\left(x_k, y_k, t_k-\Delta t_k\right)=\pol_k C.
\end{equation}
Each event contains the space-time coordinates ($x_k,y_k,t_k$) of the brightness change and its polarity $\pol_k=\{+1,-1\}$, %
with the elapsed time $\Delta t_k$ since the previous event.

\subsection{Contour Classification}
\label{sec:method:classification}

Since events occur due to brightness changes,
they do so not only at object contours
but at any edges in the image plane, such as textures,
assuming the scene brightness is constant.
Hence,
it is necessary to classify events derived from contours rather than other events for carving objects.
For the contour classification,
we utilize a convolutional neural network (CNN) encoder-decoder model (\cref{fig:classification_network})  \cite{Wang20eccv}.
As the model input,
we collect the most recent $10,000$ events
and use a voxel representation (three-dimensional tensor) by discretizing position and time.
\begin{equation}
\label{eq:eventvolume}
    \mathrm{C}_i(x,y,t)={p_ik_b(x-x_i)k_b(y-y_i)k_b(t-t_i)},
\end{equation}
where $k_b$ is a bilinear kernel, allocating event coordinates $(x,y,t)$ to each bin $(x_i,y_i,t_i)$.
The bilinear voting enables the model to learn based on
the temporal and spatial relationships between events.

The model is trained in a supervised manner,
minimizing binary cross-entropy loss
between the ground truth (GT) labels $\hat{q_i}$ from mask
and the contour inference result $f_\theta(e_i, C_i)$ for the \emph{i}-th event:
\begin{equation}
\label{eq:bce}
    \mathcal{L}_c=\frac{1}{N}\sum_{i=1}^{N}{\mathcal{L}_{bce}(f_\theta(e_i, C_i), \hat{q}_i)}.
\end{equation}
As results of the training,
as shown in \cref{fig:classification_network},
we can extract human body contours from the original event data (i.e., event stream).

\subsection{Voxel Carving}
\label{sec:method:carving}
\def\figWidth{1.0\linewidth}
\begin{figure}[t]
    \begin{center}
        {\includegraphics[trim={1cm 8cm 0.3cm 2cm},clip,width=\figWidth]{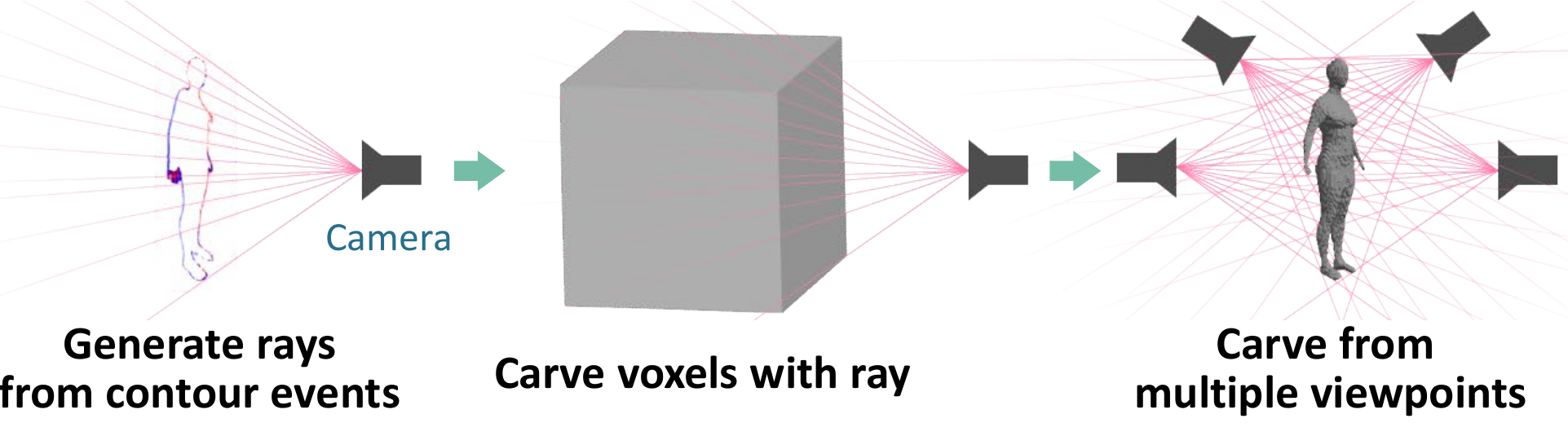}}
        \caption{Carving.
        }
        \label{fig:human_carving}
    \end{center}
\end{figure}

We carve voxels using the contour events extracted in the previous step (\cref{fig:human_carving}).
First, we convert the image coordinates $(x, y)$ 
of individual events
to the world coordinates $(X_W, Y_W, Z_W)$
via the camera coordinate system $(X_C, Y_C, Z_C)$,
using the camera intrinsic parameters $K$,
the rotation matrix $R$,
and the translation $T$ obtained from the input camera trajectory. %
\begin{equation}
\label{eq:projection}
    \begin{bmatrix} X_W \\ Y_W \\ Z_W \end{bmatrix}=R^{-1}\begin{bmatrix} X_C \\ Y_C \\ Z_C \end{bmatrix} - T
    =sR^{-1}K^{-1}\begin{bmatrix} x\\ y \\ 1 \end{bmatrix} - T.
\end{equation}
The conversion from the two-dimensional coordinates $(x,y)$
to the three-dimensional coordinates $(X_W, Y_W, Z_W)$ is a back-projection,
and the contours in the world coordinates are represented as rays that contain an unknown scale variable $s$ \eqref{eq:projection},
i.e., lines from the camera origin. 

Furthermore, we propose a ray attenuation that is inversely proportional to the distance from the camera
to mitigate pixel quantization errors.
The accuracy of the contour rays depends on 
the pixel pitch,
and the error is proportional to the distance of the camera rays
(the error increases by $n$ when the distance from the camera is increased by $n$).
We find that 
attenuating the rays
inversely proportional to the distance
enables preserving the details of the carved voxels.
Now the rays' influence $\ray'$ during carving becomes
\begin{equation}
\label{eq:decay}
    \ray'=\frac{\ray}{d+1},
\end{equation}
with the distance from the camera $d$ and the original influence $r$.
The number of rays (intensity) passing through each voxel is weighted using $\ray'$,
and voxels above a threshold are pruned.
We find that event-based carving leverages the high temporal resolution of event data, enabling smooth 3D reconstruction.

\subsection{SMPL Fitting}
\label{sec:method:regression}
\def\figWidth{1.0\linewidth}
\begin{figure}[t]
    \begin{center}
        {\includegraphics[trim={1cm 3cm 0.7cm 3cm},clip,width=\figWidth]{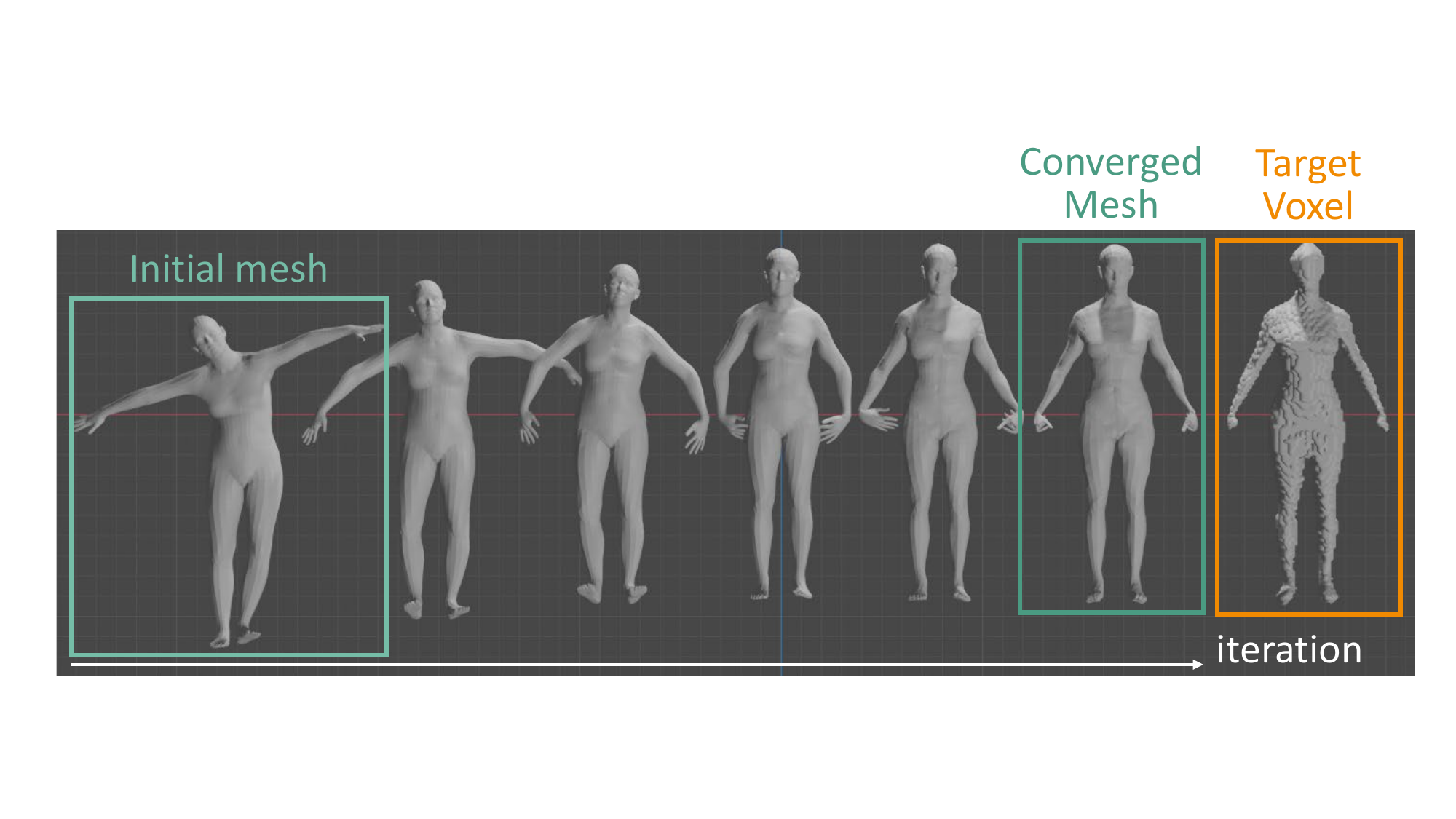}}
        \caption{SMPL fitting.
        }
        \label{fig:SMPLfitting}
    \end{center}
\end{figure}

The voxels obtained through carving do not reproduce fine structures,
such as fingertips or concave surfaces like faces or navels.
Therefore, we perform mesh reconstruction
by fitting statistical body models, SMPL \cite{SMPL:2015}.
Notice that the proposed framework can also be used for different models, such as SKEL \cite{keller2023skel} as shown in \cref{sec:result:skel}.

The SMPL model defines a function $W$ that outputs a human body mesh $M(\beta, \theta)$ with $6890$ vertices:
\begin{equation}
\begin{split}
\label{eq:SMPL}
    M(\beta, \theta) &= W(T_P(\beta, \theta), J(\beta), \theta, \mathcal{W}),\\
    T_P(\beta, \theta) &= \bar{T} + B_S(\beta) + B_P(\theta).
\end{split}
\end{equation}
The inputs of the model are
the body shape parameter $\beta\in\mathbb{R}^{10}$ (v1.0) or $\beta\in\mathbb{R}^{300}$ (v1.1)
and the pose parameters $\theta\in\mathbb{R}^{72}$ that
consists of $3$ degrees of freedom (rotational angles) for $24$ joints.
Here, $J(\beta)$ represents the joint positions considering the body shape parameters, and $\mathcal{W}$ represents the correspondence between vertices and joints.
Additionally, $T_P(\beta, \theta)$ is a human body mesh that incorporates variations in body shape and 
deformation of the flesh according to posture, 
created from the template mesh $\bar{T}$.

We fit the SMPL parameters $\theta$ and $\beta$ to the mesh that is obtained by
applying the marching cubes method \cite{lorensen1998marching} on the carved voxels.
The fitting becomes a process of
minimizing the Chamfer loss, i.e.,
\begin{equation}
\begin{split}
\label{eq:regression}
    \hat{\theta}, \hat{\beta} &= \argmin_{\theta, \beta}{(\mathcal{L}_c)},\\
    \mathcal{L}_c &= \sum_p{\min_q{||p-q||^2_2}}+\sum_q{\min_p{||p-q||^2_2}}.
\end{split}
\end{equation}
Here, $p$ and $q$ are $50,000$ points randomly sampled from the surfaces of the SMPL model and the carving result mesh, respectively,
and we calculate the sum of squared Euclidean distances between the nearest $q$ and $p$ for all pairs.
We use the Adam optimizer \cite{kingma2014adam} with a learning rate of $0.01$ and $1000$ iterations,
and finally obtain the estimation of the body parameters $(\hat{\theta}, \hat{\beta})$.

\section{Experiments}
\label{sec:experim}

In this section, we first present the dataset used in the experiments in \cref{sec:dataset},
explain the evaluation metrics and the frame-based methods used to benchmark in \cref{sec:benchmarking},
and discuss the results in \cref{sec:results} comparing other baselines.
Finally, we discuss the advantages of the proposed method under motion blur in \cref{sec:result:blur}.

\subsection{Dataset}
\label{sec:dataset}
\def\figWidth{1.0\linewidth}
\begin{figure*}[t]
    \begin{center}
        \includegraphics[trim={6cm 12cm 3cm 5cm},clip, width=\figWidth]{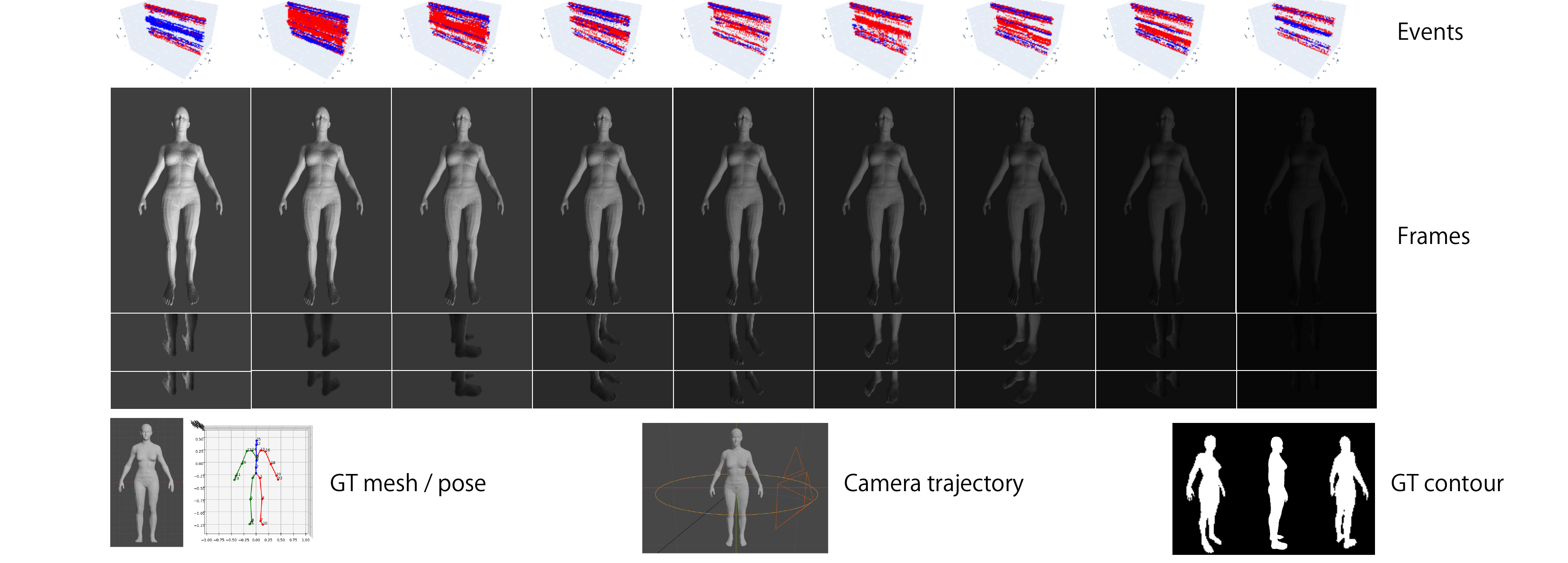}
        \caption{Dataset.}
        \label{fig:dataset_ex}
    \end{center}
\end{figure*}

The proposed problem setting of moving event cameras
for human body mesh reconstruction is,
to the best of our knowledge, the first attempt.
As there are no datasets available for such problem settings,
we create a new dataset using the event camera simulator (ESIM) \cite{Rebecq18corl}.
ESIM %
generates raw event stream 
from object 3D model,
camera intrinsic parameters,
and camera trajectory.
In our case, the human mesh is represented by a .obj file with the joint ground truth (GT) data created beforehand from the SMPL model,
and the camera trajectory follows a path circling the subject twice at a radius of $1~\si{\meter}$.
The camera is modeled as a pinhole camera,
with parameters of
$(\text{width}, \text{height}, f_x, f_y, c_x,c_y) = (640, 480, 250, 250, 320, 240)$
.
We also generate $2082$ contour GT data with timestamps for training the contour classification network,
by binarizing the depth GT from ESIM.
Additionally, we save frame images at $30$~fps
for the baselines.
Example data in the dataset are shown in \cref{fig:dataset_ex}.
In total, we prepare 27 sequences
that consist of three conditions (two different poses and one motion-blur sequence) with nine illumination settings.
Each sequence contains approximately $20$ million events and $518$ frame images.

\subsection{Evaluation Metrics and Baselines}
\label{sec:benchmarking}

\textbf{PEL-MPJPE.}
As quantitative evaluation metrics for joints,
we use the Pelvis-Aligned Mean Per Joint Position Error (PEL-MPJPE),
used in human pose estimation work \cite{zou2021eventhpe, Kocabas20cvpr}.
MPJPE represents the average 3D Euclidean distance
between the estimated joint positions and the ground truth joint positions, such as
\begin{equation}
\label{MPJPE}
    \mathbf{MPJPE}(x,\hat{x})=\frac{1}{N}\sum_{i=1}^{N}\| p_{\hat{x}}(i)-p_{x}(i)\|,
\end{equation}
where $N$ denotes the total number of joints,
$p_x$ is the joint position of the GT mesh,
and $p_{\hat{x}}$ is the estimated joint positions.
PEL-MPJPE evaluates the accuracy of relative joint positions
with respect to the root (pelvis) position after aligning them. %

\textbf{Chamfer Distance.}
For the evaluation of body meshes,
we use the Chamfer Distance (CD):
\begin{equation}
\begin{split}
\label{eq:chamferDistance}
    \mathbf{CD}(X,\hat{X}) = & \frac{1}{|X|}
    \sum_{x\in{X}}{\min_{\hat{x}\in{{\hat{X}}}}{\| x-\hat{x}\|^2_2}}
    \\ & + \frac{1}{\hat{X}}\sum_{\hat{x}\in{\hat{X}}}{\min_{x\in{X}}{\|x-\hat{x}\|^2_2}}.
\end{split}
\end{equation}
Similar to the Chamfer Loss used in \cref{sec:method:regression} \eqref{eq:regression},
CD measures the similarity of meshes based on the 3D Euclidean distance of sampled points from the mesh surfaces.
Here, $X$ is the point cloud sampled from the GT mesh,
and $\hat{X}$ is the point cloud sampled from the estimated mesh.
While in Chamfer Loss (\cref{sec:method:regression}) during fitting the body model
we sample $50$k points, %
here we sample $1$M points from the GT mesh and the recovered mesh surface for evaluation.
Both MPJPE and CD are averaged over the results from the 9 sequences with different lighting conditions in the dataset.

\textbf{Baselines}.
Existing methods of event-based human body mesh reconstruction
track moving humans with stationary event cameras
and require additional frame images.
Hence, it is challenging to directly compare them with the proposed method,
where the camera moves around stationary human bodies,
and which does not rely on frames.
Therefore, as baselines,
we use three frame-based methods:
PyMAF \cite{pymaf2021},
its extension PyMAF-X \cite{pymafx2023},
and another state-of-the-art EasyMoCap \cite{easymocap}.
PyMAF and PyMAF-X regress the SMPL parameters that fit a single RGB image,
and EasyMoCap fits the SMPL model from multiple cameras (viewpoints).
For PyMAF and PyMAF-X,
to fairly compare with the proposed method using multi-view events,
we compare the averaged outputs when using 8 frames with viewpoints changing at $45\si{\degree}$ intervals (\emph{multi-image})
and when using all $518$ frame images output at $30$~fps from ESIM (\emph{all-image}).

\subsection{Comparison with Frame-based Methods}
\label{sec:results}

\begin{table}[t]
    \centering
    \caption{Quantitative evaluation results.
    }
    \label{tab:mpjpe}
    \setlength{\tabcolsep}{3pt}
    \begin{tabular}{cccccccc}
        \hline
          & & \multicolumn{2}{c}{PEL-MPJPE [mm] $\downarrow$} & \multicolumn{2}{c}{CD [mm] $\downarrow$} \\
          & & Pose1 & Pose2 & Pose1 & Pose2 \\ \midrule
        \multicolumn{2}{c}{Ours (Event-based)} & \bnum{58.11}  & \bnum{65.64} & \bnum{7.589} & \bnum{13.36} \\\midrule
        \multicolumn{2}{c}{EasyMocap \cite{easymocap}} & 59.54  & 97.29 & 36.18 & 30.09 \\\midrule
        \multirow{3}{*}{PyMAF \cite{pymaf2021}}
        & single   & 78.40 & 234.1 & 49.83 & 239.8\\ 
        & multi & 63.90  & 185.4 & 30.19  & 193.6\\ 
        & all & 64.09  & 195.2 & 29.82  & 203.5\\\midrule
        \multirow{3}{*}{PyMAF-X \cite{pymafx2023}}
        & single & 92.10  & 182.9 & 48.68    & 199.1\\  
        & multi & 82.49  & 109.1 & 22.54 & 61.49\\
        & all   & 81.95  & 108.6  & 22.16 & 60.12\\
        
        \bottomrule

    \end{tabular}
\end{table}

\def\figWidth{0.9\linewidth}
\begin{figure}[t]
    \begin{center}
        \includegraphics[trim={1cm 5cm 3cm 5.5cm},clip,width=\figWidth]{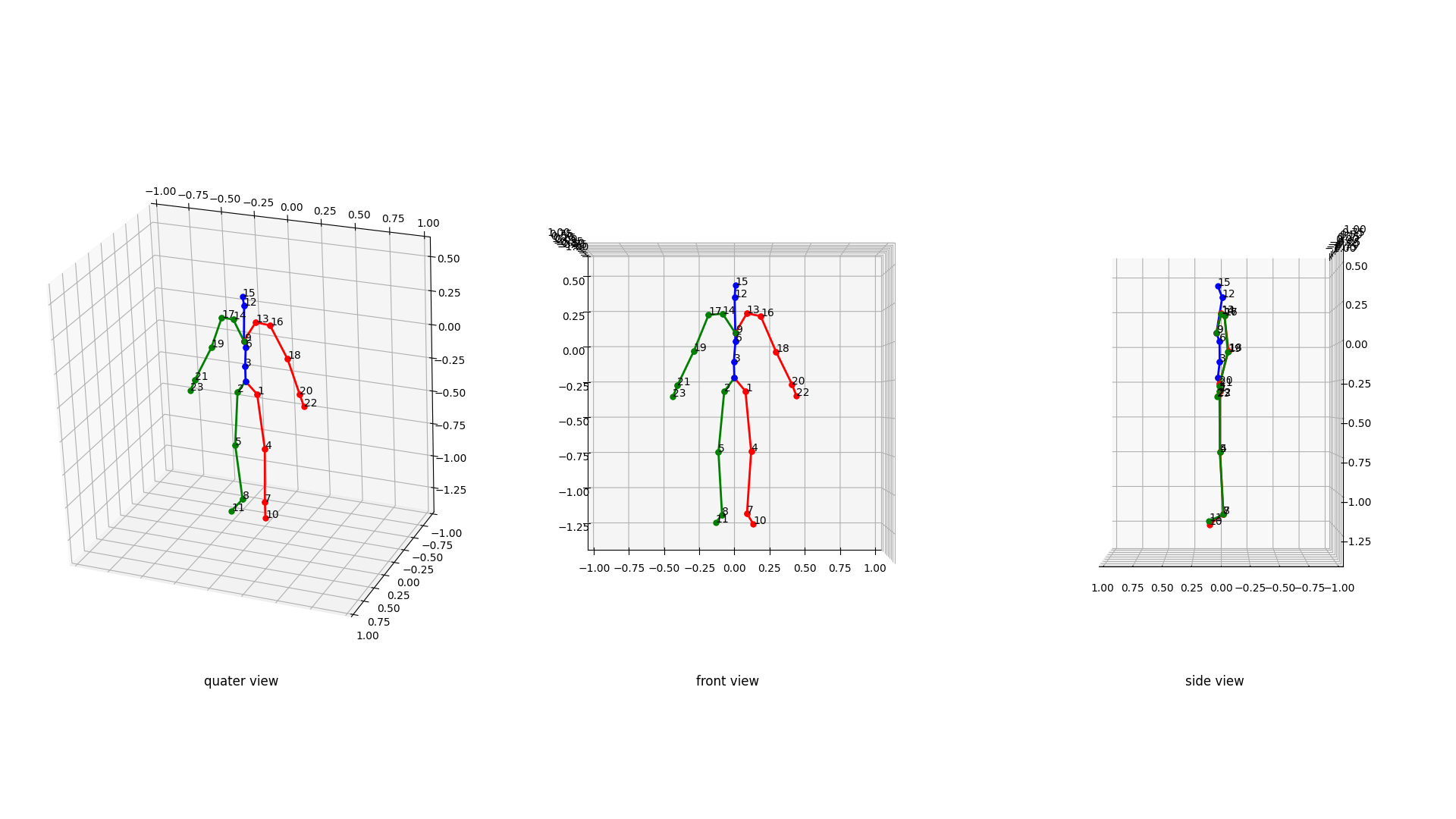} \\
        \includegraphics[trim={1cm 5cm 3cm 5.5cm},clip,width=\figWidth]{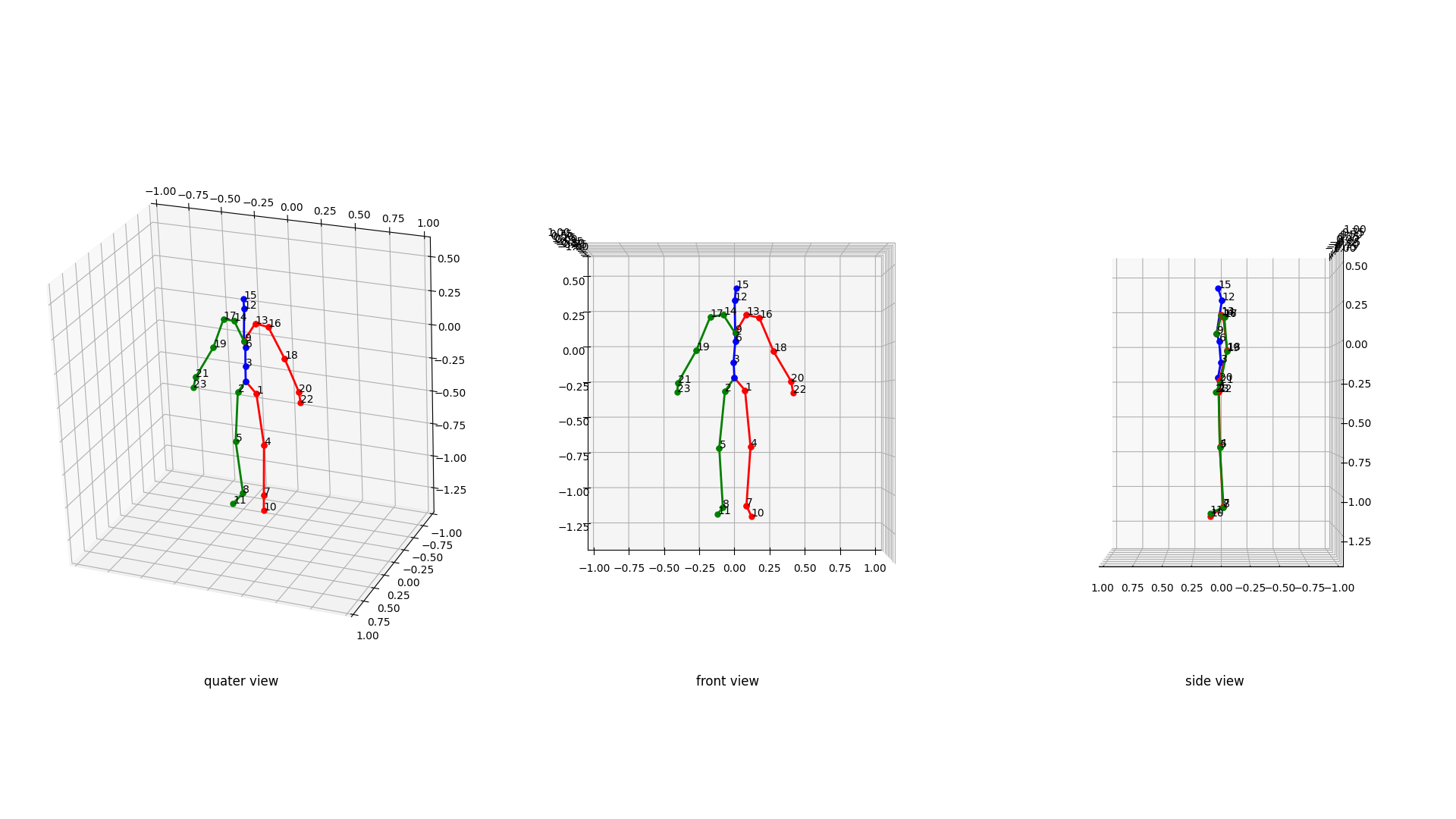}
        \caption{Qualitative results of the pose estimation for GT (top) and the proposed method (bottom).
        }
        \label{fig:skeleton_scan}
    \end{center}
\end{figure}

\Cref{tab:mpjpe} shows the quantitative results compared with the frame-based baselines.
Our proposed method achieves the smallest errors
in MPJPE and CD for both ``Pose1'' (A-pose),
and ``Pose2'' (T-like pose).
In particular,
we observe significant improvements compared with PyMAF and PyMAF-X.
Although these baselines are proposed as a single-view--estimation method,
it is remarkable that the proposed event-based approach realizes $30$--$60\%$ improvement in MPJPE and $80$--$90\%$ improvement in CD.
On the other hand, EasyMoCap %
results in competitive accuracy for ``Pose1'',
while its results for ``Pose2'' are not as good.
Nevertheless, our proposed method consistently achieves low errors ($55$--$65~\si{\milli\meter}$ for MPJPE and $5$--$15~\si{\milli\meter}$ for CD) for both poses.
These results demonstrate the effectiveness of the proposed carving and SMPL fitting combination for different poses and different illumination conditions.
\def\figWidth{1.0\linewidth}
\def\figWidth{1.0\linewidth}
\begin{figure}[t!]
    \begin{center}
        {\includegraphics[trim={15cm 2cm 16cm 0cm},clip,width=\figWidth]{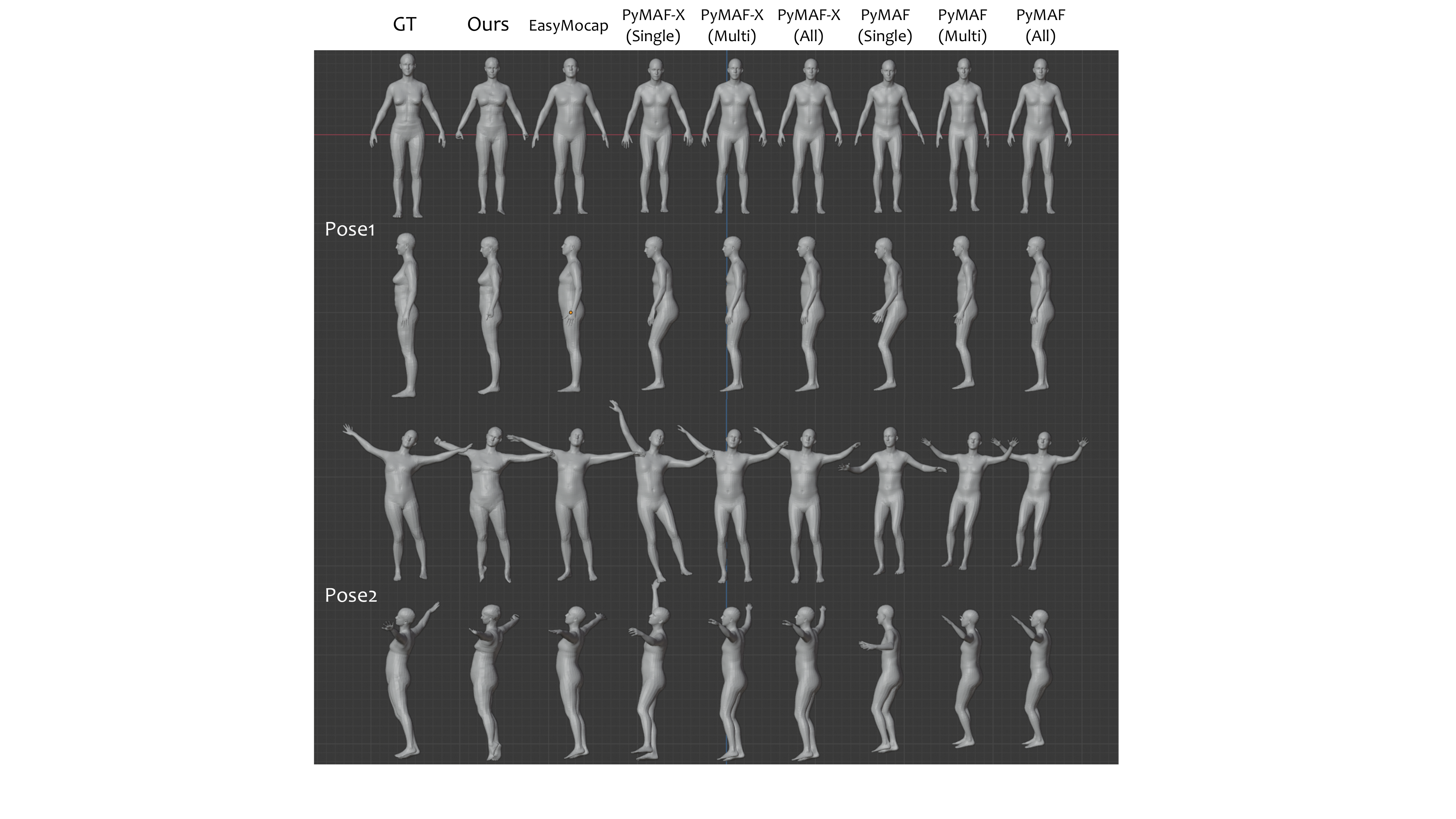}}
        \caption{Qualitative comparison of the mesh reconstruction, from the front (top) and side (bottom).
        }
        \label{fig:qualitative}
    \end{center}
\end{figure}

The qualitative results are shown in \cref{fig:skeleton_scan} (poses)
and in \cref{fig:qualitative} (meshes).
The proposed method can estimate 3D poses without major discrepancies, compared with the GT skeleton (\cref{fig:skeleton_scan}).
Furthermore,
the reconstructed meshes (\cref{fig:qualitative}) have significantly better lateral views,
compared to frame-based methods.
However, the reconstructed meshes of the proposed method
exhibit minor structural distortions in fingertips and toes,
and the overall mesh is slightly smaller.
This is attributed to missing detailed information during carving due to sampling errors and event noises.
We discuss in detail the effect of such missing structure,
together with the effect of ray attenuation in \cref{sec:result:ablation}.

\subsection{Results on Motion-Blur Sequences}
\label{sec:result:blur}

One remarkable advantage of event cameras is their minimal motion blur.
Therefore, we evaluate the accuracy by moving the camera at ten times faster,
where the frames suffer from blurry images (\cref{fig:blurred_frame}). 
\Cref{tab:blurred_quantity} reports the quantitative results.
Compared with the results without any blur (\cref{tab:mpjpe}),
existing frame-based methods have worse results in both pose and mesh accuracy due to the blur.
In contrast, the proposed method manages to mitigate the blur effect,
which clearly shows the efficacy of event cameras for such rapid motion sequences.
\begin{table}[t]
    \centering
    \caption{Results on the motion-blur sequence.
    The frame-based methods suffer from motion blur, resulting in the accuracy drop from \cref{tab:mpjpe}.}
    \label{tab:blurred_quantity}
    \begin{tabular}{ccc}\hline 
          & PEL-MPJPE [mm] $\downarrow$ & CD [mm] $\downarrow$ \\\hline
         Ours & \bnum{69.68} & \bnum{14.33} \\ 
         EasyMocap \cite{easymocap} & 223.6 & 121.0 \\
         PyMAF \cite{pymaf2021}  & 216.3  & 223.3 \\
         PyMAF-X \cite{pymafx2023} & 148.6  & 109.2 \\
         
        \hline       
    \end{tabular}
\end{table}

\def\figWidth{0.6\linewidth}
\begin{figure}[t]
    \begin{center}
        \includegraphics[width=\figWidth]{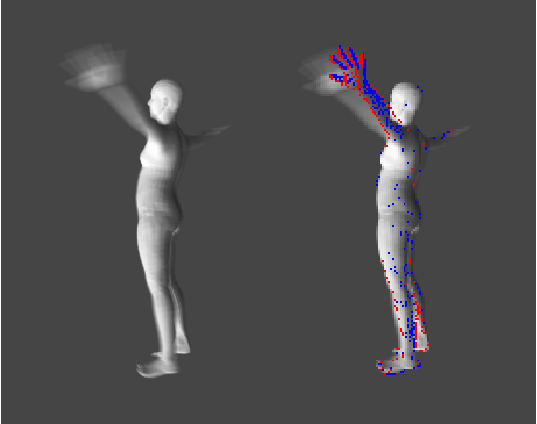}
        \caption{Visualization of the motion-blur sequence. (left) frames suffer from blur, while (right) events do not.
        }
        \label{fig:blurred_frame}
    \end{center}
\end{figure}

\section{Ablation}
\label{sec:result:ablation}

\def\figWidth{0.8\linewidth}
\begin{figure}[t]
    \begin{center}
        {\includegraphics[trim={4cm 1cm 3cm 1cm},clip,width=\figWidth]{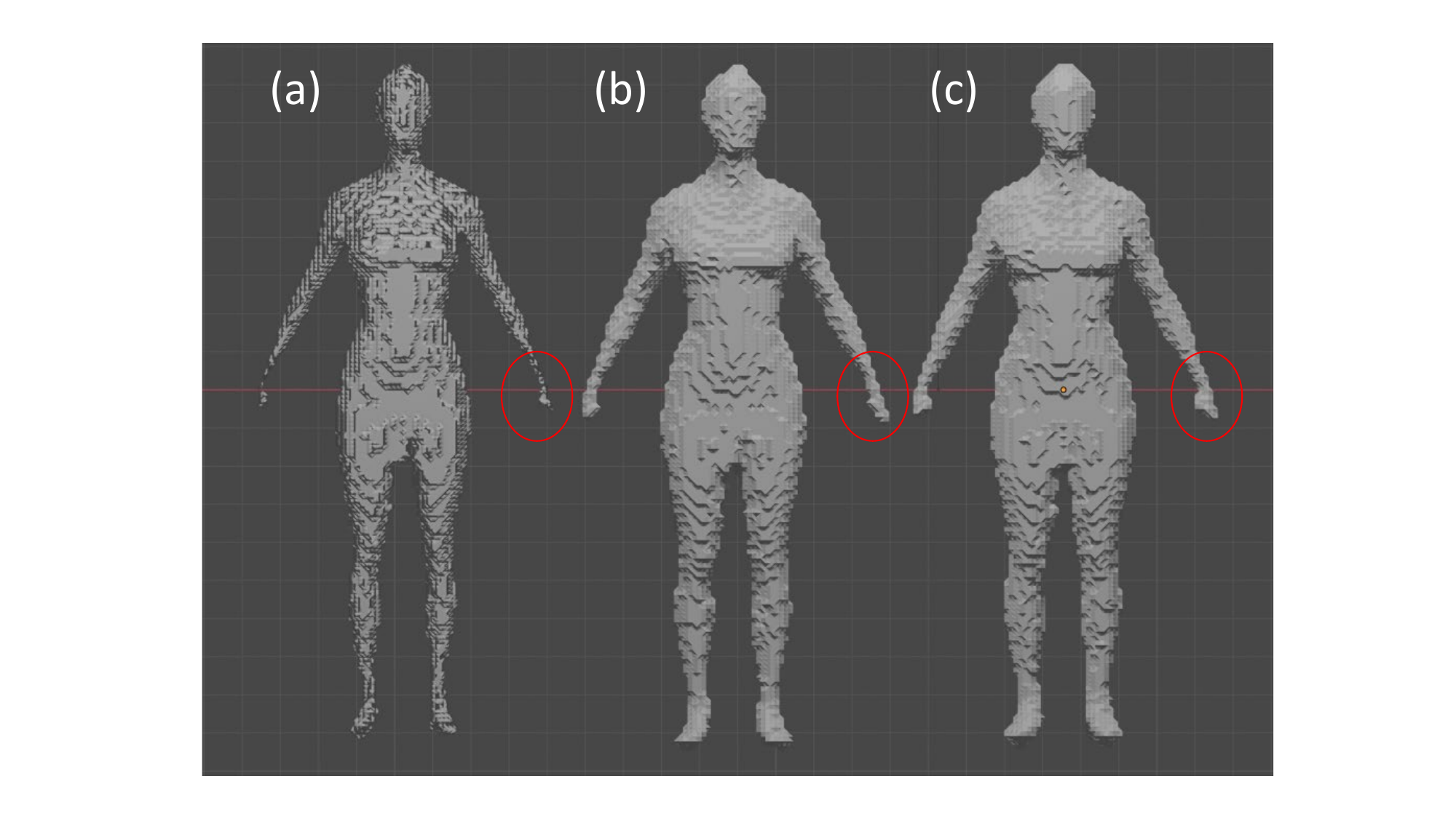}}
        \caption{Effect of the ray attenuation.
        (a) No attenuation.
        (b) Linear attenuation.
        (c) Inverse attenuation \eqref{eq:decay}.
        }
        \label{fig:decay_ablation}
    \end{center}
\end{figure}

\begin{table}[t]
    \centering
    \caption{Effect of the ray attenuation.}
    \label{tab:ablation}
    \begin{tabular}{ccc}\hline
         & PEL-MPJPE [mm] $\downarrow$ & CD [mm] $\downarrow$ \\\hline
         No attenuation & 56.90 & 144.6 \\ 
         Linear & 38.75  & 7.601 \\
         Inverse \eqref{eq:decay} & \bnum{38.24}  & \bnum{6.692}\\
        \hline       
    \end{tabular}
\end{table}

\subsection{Effect of Ray Attenuation}
\label{sec:result:decay}

To validate the effectiveness of the proposed ray attenuation \eqref{eq:decay},
we compare the voxels of carving results with
(a) no attenuation: $\ray' = \ray$,
(b) linear: $\ray'= \max{(\ray - d)}$,
and (c) inverse \eqref{eq:decay}.
\Cref{fig:decay_ablation} shows the qualitative comparison among the three.
In cases with ray attenuation (b),(c), carving loss of details
in fingertips and toes is mitigated as expected.
Quantitative evaluations of SMPL fitting on these voxels are presented in \cref{tab:ablation},
demonstrating the efficacy of the proposed decay
for both pose estimation and mesh reconstruction accuracies.

\subsection{Results on Other Parametric Model}
\label{sec:result:skel}

\def\figWidth{0.75\linewidth}
\begin{figure}[t]
    \begin{center}
        {\includegraphics[trim={15cm 5cm 15cm 5cm},clip,width=\figWidth]{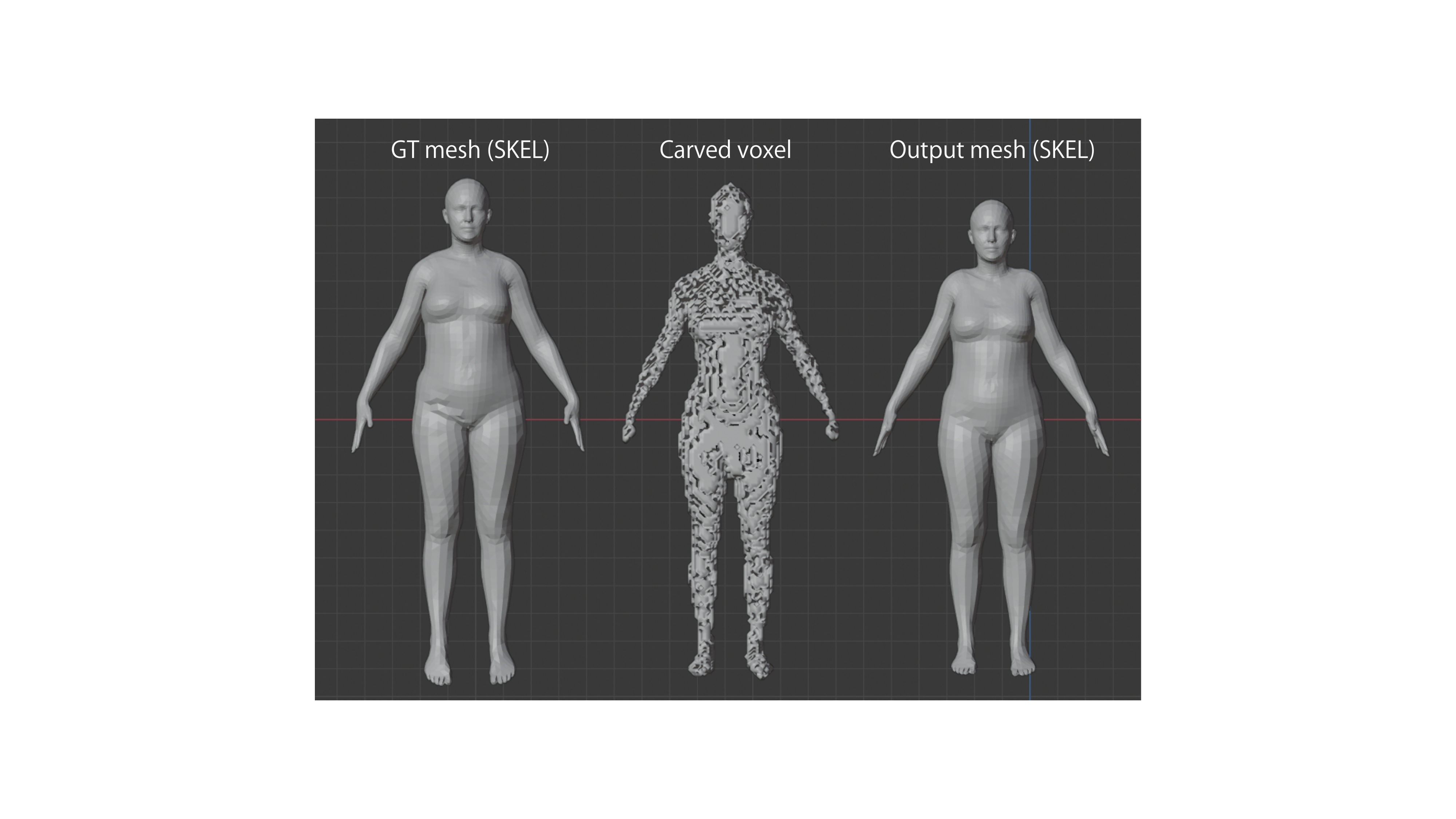}} \\
        \caption{Qualitative results on the SKEL\cite{keller2023skel} model.
        }
        \label{fig:skel_qualitative}
    \end{center}
\end{figure}

The proposed method is not limited to fitting the SMPL model.
To this end, 
we conduct experiments using SKEL \cite{keller2023skel}
for ground truth mesh and fitting.
As shown in \cref{fig:skel_qualitative}, 
the estimated meshes look reasonably similar to the GT,
with the low error of MPJPE: $54.80$ and CD: $7.943$.

\subsection{Comparison with GT masks}
\label{sec:result:gtmask}
In addition to the frame-based baselines,
we conduct another experiment that confirms the advantage of high temporal resolution of event cameras.
To this end, we use the GT contour (\cref{fig:classification_network,fig:dataset_ex}) at $30$~fps, replacing the contour classification network in the proposed pipeline.
The contour edges obtained from GT are used for the same carving and SMPL fitting steps.
As results, the mesh estimated from the GT contour
achieves slightly better pose estimation (MPJPE: $40.54$),
while the proposed method achieves better mesh reconstruction (CD: $14.74$).
The quality of the carved voxel is also confirmed in \cref{fig:gt_mask}.
\def\figWidth{0.8\linewidth}
\begin{figure}[t]
    \begin{center}
        {\includegraphics[trim={4cm 1cm 4cm 0cm},clip,width=\figWidth]{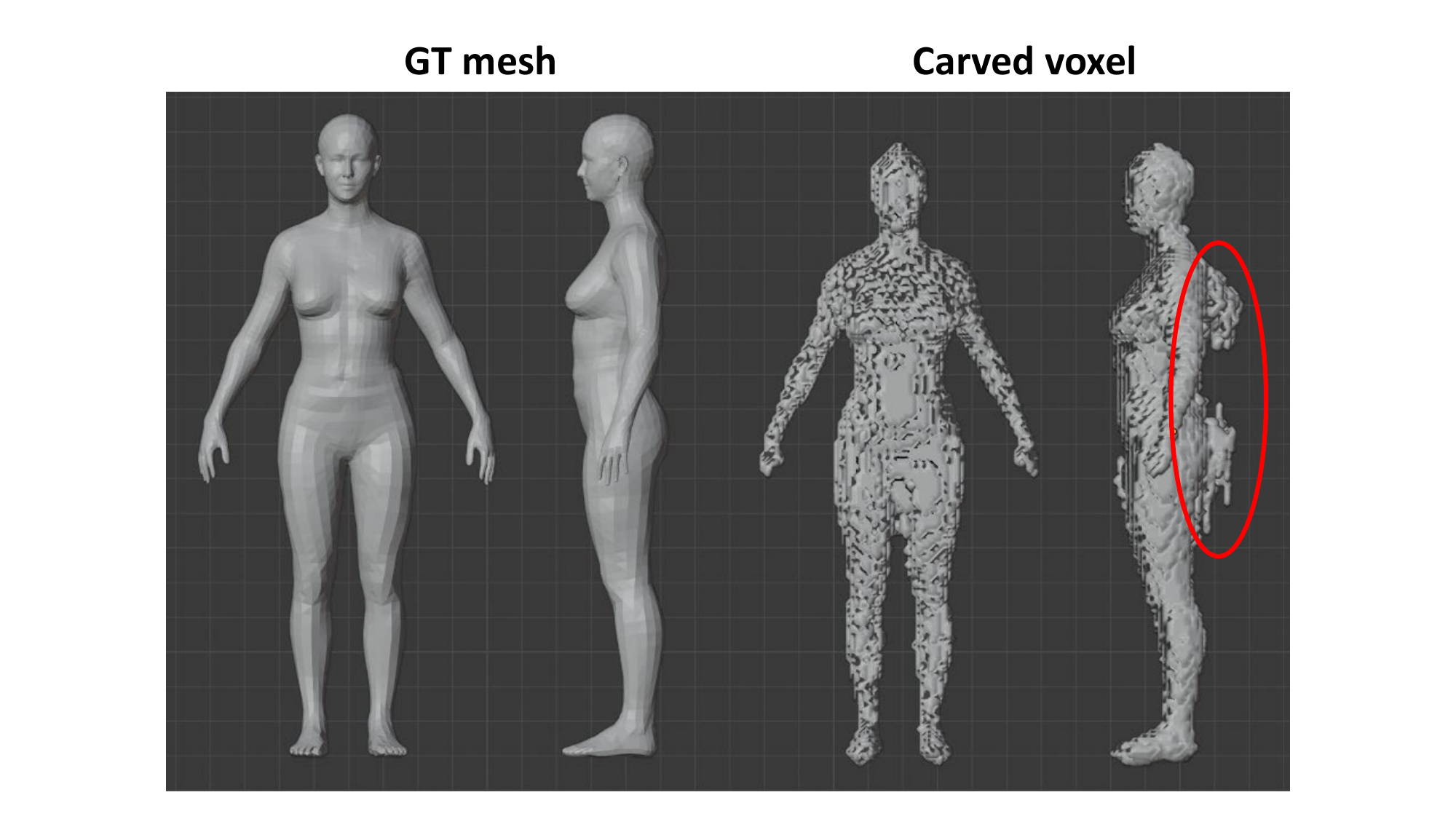}} \\
        \caption{Qualitative results of carving with GT masks, from the front and side.
        }
        \label{fig:gt_mask}
    \end{center}
\end{figure}

\section{Sensitivity Study}
\label{sec:result:sensitivity}

Finally, we conduct sensitivity studies for the proposed method.
Here, we compare the results when
($i$) changing the dimensions of body shape parameters $\beta$ in SMPL ($300$, $10$),
and ($ii$) changing the voxel size (i.e., the numbers of voxel divisions) during carving ($512$, $256$, $128$).
Larger values of $\beta$ can fit finer details,
however, they are prone to overfitting on excessively carved voxels.
Increasing the number of voxels allows for better representation of detailed shapes,
while it requires more rays, which can result in uncarved voxels remaining.
The quantitative results in \cref{tab:sensitivity} show
that both joint estimation accuracy and mesh reconstruction accuracy surpass when $\beta\in\mathbb{R}^{10}$ compared to $\beta\in\mathbb{R}^{300}$.
Being consistent with the qualitative results (\cref{fig:result_sensitivity}),
it suggests that expressive $\beta\in\mathbb{R}^{300}$ fits meshes to excessively trimmed terminal parts such as wrists,
leading to increased deviation from GT (i.e., overfitting).
Moreover, increasing the number of voxels improves estimation accuracy.
However,
mesh reconstruction tends to fail ($\mathrm{voxel\ size}=512$ in \cref{tab:sensitivity})
due to insufficient rays. %
Hence, in the experimental setup tested,
$\beta\in\mathbb{R}^{10}, \mathrm{voxel\ size}=256$ is validated as the optimal hyperparameters.

\begin{table}[t]
    \centering
    \caption{Sensitivity study on the dimension of $\beta$ and the voxel size.
    $\ast$ denotes the average without outliers (when the reconstruction fails),
    and $\dagger$ denotes the average with such outliers.}
    \label{tab:sensitivity}
    \begin{tabular}{cccc}
        \hline
        $\beta$ dim   
        & Voxel size & PEL-MPJPE [mm] $\downarrow$ & CD [mm] $\downarrow$ \\\hline
        \multirow{4}{*}{300}
         &512$\ast$ & 61.80 & 7.894 \\ 
         &512$\dagger$ & 114.4 & 81.83 \\ 
         &256   & 71.90  & 10.38 \\
         &128 & 142.0  & 36.97\\\midrule
         \multirow{4}{*}{10}
         &512$\ast$ & 55.42 & 6.800 \\
         &512$\dagger$ & 111.1 & 80.93 \\
         & 256  & \bnum{58.11}  & \bnum{7.589}\\
         & 128 & 151.3  & 36.38\\
        \hline       
    \end{tabular}
\end{table}

\def\figWidth{0.7\linewidth}
\begin{figure}[h]
    \begin{center}
        {\includegraphics[trim={10.5cm 0.5cm 12.5cm 0.5cm},clip,width=\figWidth]{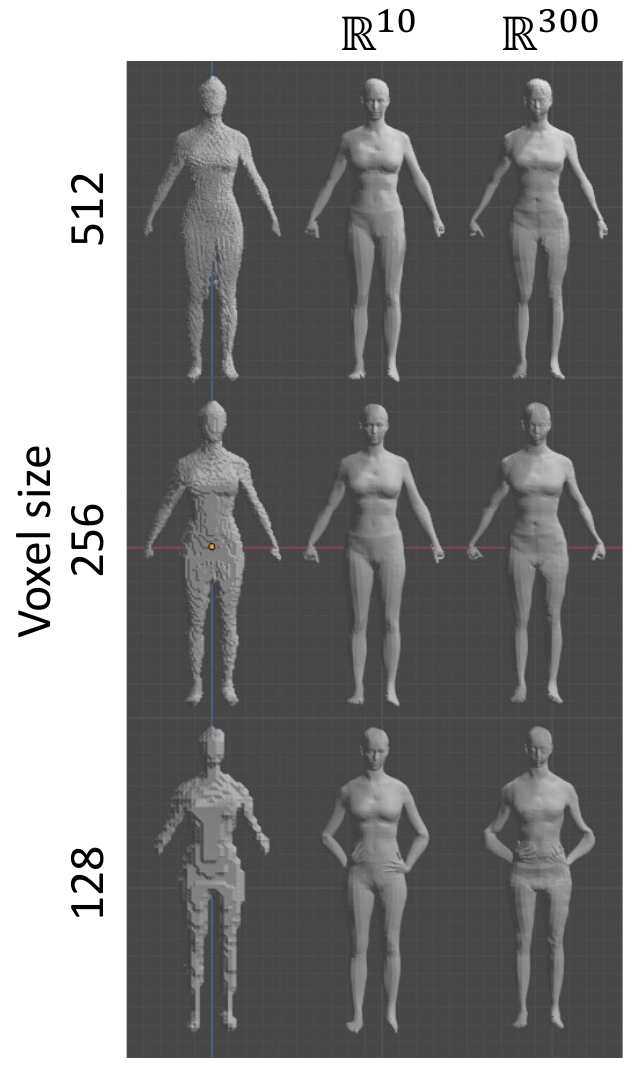}}
        \caption{Qualitative result of sensitivity study.
        }
        \label{fig:result_sensitivity}
    \end{center}
\end{figure}

\section{Limitations}
\label{sec:limitation}

In this work,
we propose the first-of-its-kind method to estimate static human pose and mesh from only event stream,
using a moving event camera.
The proposed method %
combines the classical idea of carving and fitting statistical models of the human body
with the ray attenuation to preserve fine structures of the subject,
and utilizes the high temporal information of event data.
The experimental results show that
($i$) the proposed event-based method achieves better accuracy than frame-based baselines both in pose and mesh estimation,
($ii$) the ray attenuation is effective for mesh recovery with high-frequency details,
($iii$) the high temporal resolution contributes to the precise carving,
and ($iv$) the proposed method is robust against motion-blur scenes.
We hope this work will serve as a baseline to be extended to various interesting directions in the future as followings.

The proposed method needs the camera trajectory,
necessitating ego-motion estimation for the real-world applications.
Since the scene (body) is static,
one interesting direction could be the combination with SLAM (simultaneous localization and mapping) methods.
Also, the contour classification utilizes supervised learning,
while the other steps (carving and SMPL fitting) are optimization method.
Thus,
the contour classification performance may degrade for textures (e.g., clothes) or different poses that are not included in the training data.
Creating a large-scale human body dataset with diverse textures and poses could be another important direction.
Finally, the proposed method still may miss some fine structures as discussed in \cref{sec:result:sensitivity},
which can be attributed to the limitation of the carving.
Recently there have been many work for spatial understanding,
such as Neural Radiance Field (NeRF) \cite{mildenhall2021nerf}
and Gaussian Splatting \cite{kerbl3Dgaussians},
including dynamic scenes.
Combining these ideas into the human scanning system could be a unique contribution in the future.

{
    \small
    % \bibliographystyle{ieeenat_fullname}
    % \bibliography{all}

\begin{thebibliography}{55}
\providecommand{\natexlab}[1]{#1}
\providecommand{\url}[1]{\texttt{#1}}
\expandafter\ifx\csname urlstyle\endcsname\relax
  \providecommand{\doi}[1]{doi: #1}\else
  \providecommand{\doi}{doi: \begingroup \urlstyle{rm}\Url}\fi

\bibitem[eas(2021)]{easymocap}
Easymocap - make human motion capture easier.
\newblock Github, 2021.

\bibitem[Angelopoulos et~al.(2020)Angelopoulos, Martel, Kohli, Conradt, and
  Wetzstein]{Angelopoulos20tvcg}
Anastasios~N Angelopoulos, Julien~NP Martel, Amit~PS Kohli, Jorg Conradt, and
  Gordon Wetzstein.
\newblock Event based, near eye gaze tracking beyond 10,000 hz.
\newblock \emph{{IEEE} Trans. Vis. Comput. Graphics}, 2020.

\bibitem[Benosman et~al.(2012)Benosman, Ieng, Clercq, Bartolozzi, and
  Srinivasan]{Benosman12nn}
Ryad Benosman, Sio-Hoi Ieng, Charles Clercq, Chiara Bartolozzi, and Mandyam
  Srinivasan.
\newblock Asynchronous frameless event-based optical flow.
\newblock \emph{Neural Netw.}, 27:\penalty0 32--37, 2012.

\bibitem[Calabrese et~al.(2019)Calabrese, Taverni, Easthope, Skriabine,
  Corradi, Longinotti, Eng, and Delbruck]{Calabrese19cvprw}
Enrico Calabrese, Gemma Taverni, Christopher~Awai Easthope, Sophie Skriabine,
  Federico Corradi, Luca Longinotti, Kynan Eng, and Tobi Delbruck.
\newblock {DHP19}: Dynamic vision sensor {3D} human pose dataset.
\newblock In \emph{{IEEE} Conf. Comput. Vis. Pattern Recog. Workshops (CVPRW)},
  2019.

\bibitem[Chen and Ramanan(2017)]{chen20173d}
Ching-Hang Chen and Deva Ramanan.
\newblock 3d human pose estimation= 2d pose estimation+ matching.
\newblock In \emph{Proceedings of the IEEE conference on computer vision and
  pattern recognition}, pages 7035--7043, 2017.

\bibitem[Chen et~al.(2022)Chen, Shi, Ye, Yang, Sun, and
  Wang]{chen2022efficient}
Jiaan Chen, Hao Shi, Yaozu Ye, Kailun Yang, Lei Sun, and Kaiwei Wang.
\newblock Efficient human pose estimation via 3d event point cloud.
\newblock In \emph{2022 International Conference on 3D Vision (3DV)}, pages
  1--10. IEEE, 2022.

\bibitem[Dong et~al.(2019)Dong, Jiang, Huang, Bao, and Zhou]{dong2019fast}
Junting Dong, Wen Jiang, Qixing Huang, Hujun Bao, and Xiaowei Zhou.
\newblock Fast and robust multi-person 3d pose estimation from multiple views.
\newblock In \emph{Proceedings of the IEEE/CVF conference on computer vision
  and pattern recognition}, pages 7792--7801, 2019.

\bibitem[Finateu et~al.(2020)Finateu, Niwa, Matolin, Tsuchimoto, Mascheroni,
  Reynaud, Mostafalu, Brady, Chotard, LeGoff, Takahashi, Wakabayashi, Oike, and
  Posch]{Finateu20isscc}
Thomas Finateu, Atsumi Niwa, Daniel Matolin, Koya Tsuchimoto, Andrea
  Mascheroni, Etienne Reynaud, Pooria Mostafalu, Frederick Brady, Ludovic
  Chotard, Florian LeGoff, Hirotsugu Takahashi, Hayato Wakabayashi, Yusuke
  Oike, and Christoph Posch.
\newblock A 1280x720 back-illuminated stacked temporal contrast event-based
  vision sensor with 4.86$\mu$m pixels, 1.066{G}eps readout, programmable
  event-rate controller and compressive data-formatting pipeline.
\newblock In \emph{{IEEE} Intl. Solid-State Circuits Conf. (ISSCC)}, pages
  112--114, 2020.

\bibitem[Gallego et~al.(2018)Gallego, Rebecq, and Scaramuzza]{Gallego18cvpr}
Guillermo Gallego, Henri Rebecq, and Davide Scaramuzza.
\newblock A unifying contrast maximization framework for event cameras, with
  applications to motion, depth, and optical flow estimation.
\newblock In \emph{{IEEE} Conf. Comput. Vis. Pattern Recog. (CVPR)}, pages
  3867--3876, 2018.

\bibitem[Gallego et~al.(2022)Gallego, Delbruck, Orchard, Bartolozzi, Taba,
  Censi, Leutenegger, Davison, Conradt, Daniilidis, and
  Scaramuzza]{Gallego20pami}
Guillermo Gallego, Tobi Delbruck, Garrick Orchard, Chiara Bartolozzi, Brian
  Taba, Andrea Censi, Stefan Leutenegger, Andrew Davison, J{\"o}rg Conradt,
  Kostas Daniilidis, and Davide Scaramuzza.
\newblock Event-based vision: A survey.
\newblock \emph{{IEEE} Trans. Pattern Anal. Mach. Intell.}, 44\penalty0
  (1):\penalty0 154--180, 2022.

\bibitem[Glover and Bartolozzi(2016)]{Glover16iros}
Arren Glover and Chiara Bartolozzi.
\newblock Event-driven ball detection and gaze fixation in clutter.
\newblock In \emph{IEEE/RSJ Int. Conf. Intell. Robot. Syst. (IROS)}, pages
  2203--2208, 2016.

\bibitem[Guo and Gallego(2024)]{Guo24tro}
Shuang Guo and Guillermo Gallego.
\newblock Cmax-slam: Event-based rotational-motion bundle adjustment and slam
  system using contrast maximization.
\newblock \emph{{IEEE} Trans. Robot.}, pages 1--20, 2024.

\bibitem[Jiang et~al.(2024)Jiang, Zhou, Wang, Deng, Xu, and
  Shi]{jiang2024complementing}
Jianping Jiang, Xinyu Zhou, Bingxuan Wang, Xiaoming Deng, Chao Xu, and Boxin
  Shi.
\newblock Complementing event streams and rgb frames for hand mesh
  reconstruction.
\newblock In \emph{Proceedings of the IEEE conference on computer vision and
  pattern recognition}, 2024.

\bibitem[Kanazawa et~al.(2018)Kanazawa, Black, Jacobs, and
  Malik]{hmrKanazawa17}
Angjoo Kanazawa, Michael~J. Black, David~W. Jacobs, and Jitendra Malik.
\newblock End-to-end recovery of human shape and pose.
\newblock In \emph{Computer Vision and Pattern Recognition (CVPR)}, 2018.

\bibitem[Keller et~al.(2023)Keller, Werling, Shin, Delp, Pujades, C.~Karen, and
  Black]{keller2023skel}
Marilyn Keller, Keenon Werling, Soyong Shin, Scott Delp, Sergi Pujades, Liu
  C.~Karen, and Michael~J. Black.
\newblock From skin to skeleton: Towards biomechanically accurate 3d digital
  humans.
\newblock In \emph{ACM ToG, Proc.~SIGGRAPH Asia}, 2023.

\bibitem[Kerbl et~al.(2023)Kerbl, Kopanas, Leimk{\"u}hler, and
  Drettakis]{kerbl3Dgaussians}
Bernhard Kerbl, Georgios Kopanas, Thomas Leimk{\"u}hler, and George Drettakis.
\newblock 3d gaussian splatting for real-time radiance field rendering.
\newblock \emph{ACM Transactions on Graphics}, 42\penalty0 (4), 2023.

\bibitem[Kim et~al.(2014)Kim, Handa, Benosman, Ieng, and Davison]{Kim14bmvc}
Hanme Kim, Ankur Handa, Ryad Benosman, Sio-Hoi Ieng, and Andrew~J. Davison.
\newblock Simultaneous mosaicing and tracking with an event camera.
\newblock In \emph{British Mach. Vis. Conf. (BMVC)}, 2014.

\bibitem[Kim et~al.(2016)Kim, Leutenegger, and Davison]{Kim16eccv}
Hanme Kim, Stefan Leutenegger, and Andrew~J. Davison.
\newblock Real-time {3D} reconstruction and 6-{DoF} tracking with an event
  camera.
\newblock In \emph{Eur. Conf. Comput. Vis. (ECCV)}, pages 349--364, 2016.

\bibitem[Kingma and Ba(2014)]{kingma2014adam}
Diederik~P Kingma and Jimmy Ba.
\newblock Adam: A method for stochastic optimization.
\newblock \emph{arXiv preprint arXiv:1412.6980}, 2014.

\bibitem[Kocabas et~al.(2020)Kocabas, Athanasiou, and Black]{Kocabas20cvpr}
Muhammed Kocabas, Nikos Athanasiou, and Michael~J Black.
\newblock Vibe: Video inference for human body pose and shape estimation.
\newblock In \emph{{IEEE} Conf. Comput. Vis. Pattern Recog. (CVPR)}, pages
  5253--5263, 2020.

\bibitem[Lagorce et~al.(2017)Lagorce, Orchard, Gallupi, Shi, and
  Benosman]{Lagorce17pami}
Xavier Lagorce, Garrick Orchard, Francesco Gallupi, Bertram~E. Shi, and Ryad
  Benosman.
\newblock {HOTS}: A hierarchy of event-based time-surfaces for pattern
  recognition.
\newblock \emph{{IEEE} Trans. Pattern Anal. Mach. Intell.}, 39\penalty0
  (7):\penalty0 1346--1359, 2017.

\bibitem[Laurentini(1994)]{visualhull}
A. Laurentini.
\newblock The visual hull concept for silhouette-based image understanding.
\newblock \emph{IEEE Trans. Pattern Anal. Mach. Intell.}, 16\penalty0
  (2):\penalty0 150–162, 1994.

\bibitem[Li and Lee(2019)]{li2019generating}
Chen Li and Gim~Hee Lee.
\newblock Generating multiple hypotheses for 3d human pose estimation with
  mixture density network.
\newblock In \emph{Proceedings of the IEEE/CVF conference on computer vision
  and pattern recognition}, pages 9887--9895, 2019.

\bibitem[Li et~al.(2015)Li, Zhang, and Chan]{li2015maximum}
Sijin Li, Weichen Zhang, and Antoni~B Chan.
\newblock Maximum-margin structured learning with deep networks for 3d human
  pose estimation.
\newblock In \emph{Proceedings of the IEEE international conference on computer
  vision}, pages 2848--2856, 2015.

\bibitem[Lichtsteiner et~al.(2008)Lichtsteiner, Posch, and
  Delbruck]{Lichtsteiner08ssc}
Patrick Lichtsteiner, Christoph Posch, and Tobi Delbruck.
\newblock {A 128$\times$128 120 dB 15 $\mu$s latency asynchronous temporal
  contrast vision sensor}.
\newblock \emph{{IEEE} J. Solid-State Circuits}, 43\penalty0 (2):\penalty0
  566--576, 2008.

\bibitem[Lin et~al.(2020)Lin, Zhang, Pan, Jiang, Zou, Wang, Chen, and
  Ren]{Lin20eccv}
Songnan Lin, Jiawei Zhang, Jinshan Pan, Zhe Jiang, Dongqing Zou, Yongtian Wang,
  Jing Chen, and Jimmy Ren.
\newblock Learning event-driven video deblurring and interpolation.
\newblock In \emph{Eur. Conf. Comput. Vis. (ECCV)}, pages 695--710, 2020.

\bibitem[Loper et~al.(2015)Loper, Mahmood, Romero, Pons-Moll, and
  Black]{SMPL:2015}
Matthew Loper, Naureen Mahmood, Javier Romero, Gerard Pons-Moll, and Michael~J.
  Black.
\newblock {SMPL}: A skinned multi-person linear model.
\newblock \emph{ACM Trans. Graphics (Proc. SIGGRAPH Asia)}, 34\penalty0
  (6):\penalty0 248:1--248:16, 2015.

\bibitem[Lorensen and Cline(1998)]{lorensen1998marching}
William~E Lorensen and Harvey~E Cline.
\newblock Marching cubes: A high resolution 3d surface construction algorithm.
\newblock In \emph{Seminal graphics: pioneering efforts that shaped the field},
  pages 347--353, 1998.

\bibitem[Ma et~al.(2020)Ma, Yang, Ranjan, Pujades, Pons-Moll, Tang, and
  Black]{Ma20cvpr}
Qianli Ma, Jinlong Yang, Anurag Ranjan, Sergi Pujades, Gerard Pons-Moll, Siyu
  Tang, and Michael~J. Black.
\newblock Learning to dress 3d people in generative clothing.
\newblock In \emph{{IEEE} Conf. Comput. Vis. Pattern Recog. (CVPR)}, 2020.

\bibitem[Mildenhall et~al.(2021)Mildenhall, Srinivasan, Tancik, Barron,
  Ramamoorthi, and Ng]{mildenhall2021nerf}
Ben Mildenhall, Pratul~P Srinivasan, Matthew Tancik, Jonathan~T Barron, Ravi
  Ramamoorthi, and Ren Ng.
\newblock Nerf: Representing scenes as neural radiance fields for view
  synthesis.
\newblock \emph{Communications of the ACM}, 65\penalty0 (1):\penalty0 99--106,
  2021.

\bibitem[Mitrokhin et~al.(2018)Mitrokhin, Fermuller, Parameshwara, and
  Aloimonos]{Mitrokhin18iros}
Anton Mitrokhin, Cornelia Fermuller, Chethan Parameshwara, and Yiannis
  Aloimonos.
\newblock Event-based moving object detection and tracking.
\newblock In \emph{IEEE/RSJ Int. Conf. Intell. Robot. Syst. (IROS)}, pages
  1--9, 2018.

\bibitem[Moon et~al.(2018)Moon, Chang, and Lee]{moon2018v2v}
Gyeongsik Moon, Ju~Yong Chang, and Kyoung~Mu Lee.
\newblock V2v-posenet: Voxel-to-voxel prediction network for accurate 3d hand
  and human pose estimation from a single depth map.
\newblock In \emph{Proceedings of the IEEE conference on computer vision and
  pattern Recognition}, pages 5079--5088, 2018.

\bibitem[Mueggler et~al.(2017)Mueggler, Bartolozzi, and
  Scaramuzza]{Mueggler17bmvc}
Elias Mueggler, Chiara Bartolozzi, and Davide Scaramuzza.
\newblock Fast event-based corner detection.
\newblock In \emph{British Mach. Vis. Conf. (BMVC)}, 2017.

\bibitem[Muglikar et~al.(2021)Muglikar, Gallego, and
  Scaramuzza]{Muglikar21threedv}
Manasi Muglikar, Guillermo Gallego, and Davide Scaramuzza.
\newblock {ESL}: Event-based structure light.
\newblock In \emph{Int. Conf. 3D Vision (3DV)}, pages 1165--1174, 2021.

\bibitem[Park et~al.(2016)Park, Hwang, and Kwak]{park20163d}
Sungheon Park, Jihye Hwang, and Nojun Kwak.
\newblock 3d human pose estimation using convolutional neural networks with 2d
  pose information.
\newblock In \emph{Computer Vision--ECCV 2016 Workshops: Amsterdam, The
  Netherlands, October 8-10 and 15-16, 2016, Proceedings, Part III 14}, pages
  156--169. Springer, 2016.

\bibitem[Pavlakos et~al.(2017)Pavlakos, Zhou, Derpanis, and
  Daniilidis]{pavlakos2017coarse}
Georgios Pavlakos, Xiaowei Zhou, Konstantinos~G Derpanis, and Kostas
  Daniilidis.
\newblock Coarse-to-fine volumetric prediction for single-image 3d human pose.
\newblock In \emph{Proceedings of the IEEE conference on computer vision and
  pattern recognition}, pages 7025--7034, 2017.

\bibitem[Pavlakos et~al.(2019)Pavlakos, Choutas, Ghorbani, Bolkart, Osman,
  Tzionas, and Black]{SMPL-X:2019}
Georgios Pavlakos, Vasileios Choutas, Nima Ghorbani, Timo Bolkart, Ahmed A.~A.
  Osman, Dimitrios Tzionas, and Michael~J. Black.
\newblock Expressive body capture: {3D} hands, face, and body from a single
  image.
\newblock In \emph{Proceedings IEEE Conf. on Computer Vision and Pattern
  Recognition (CVPR)}, pages 10975--10985, 2019.

\bibitem[Rebecq et~al.(2018)Rebecq, Gehrig, and Scaramuzza]{Rebecq18corl}
Henri Rebecq, Daniel Gehrig, and Davide Scaramuzza.
\newblock {ESIM}: an open event camera simulator.
\newblock In \emph{Conf. on Robotics Learning (CoRL)}, pages 969--982. PMLR,
  2018.

\bibitem[Scarpellini et~al.(2021)Scarpellini, Morerio, and
  Del~Bue]{scarpellini2021lifting}
Gianluca Scarpellini, Pietro Morerio, and Alessio Del~Bue.
\newblock Lifting monocular events to 3d human poses.
\newblock In \emph{Proceedings of the IEEE/CVF Conference on Computer Vision
  and Pattern Recognition}, pages 1358--1368, 2021.

\bibitem[Shiba et~al.(2022{\natexlab{a}})Shiba, Aoki, and Gallego]{Shiba22aisy}
Shintaro Shiba, Yoshimitsu Aoki, and Guillermo Gallego.
\newblock A fast geometric regularizer to mitigate event collapse in the
  contrast maximization framework.
\newblock \emph{Adv. Intell. Syst.}, page 2200251, 2022{\natexlab{a}}.

\bibitem[Shiba et~al.(2022{\natexlab{b}})Shiba, Aoki, and Gallego]{Shiba22eccv}
Shintaro Shiba, Yoshimitsu Aoki, and Guillermo Gallego.
\newblock Secrets of event-based optical flow.
\newblock In \emph{Eur. Conf. Comput. Vis. (ECCV)}, pages 628--645,
  2022{\natexlab{b}}.

\bibitem[Shiba et~al.(2023{\natexlab{a}})Shiba, Aoki, and Gallego]{Shiba23spl}
Shintaro Shiba, Yoshimitsu Aoki, and Guillermo Gallego.
\newblock Fast event-based optical flow estimation by triplet matching.
\newblock \emph{{IEEE} Signal Process. Lett.}, pages 1--5, 2023{\natexlab{a}}.

\bibitem[Shiba et~al.(2023{\natexlab{b}})Shiba, Hamann, Aoki, and
  Gallego]{Shiba23pami}
Shintaro Shiba, Friedhelm Hamann, Yoshimitsu Aoki, and Guillermo Gallego.
\newblock Event-based background-oriented schlieren.
\newblock \emph{{IEEE} Trans. Pattern Anal. Mach. Intell.}, 2023{\natexlab{b}}.

\bibitem[Shuai et~al.(2022)Shuai, Geng, Fang, Peng, Shen, Zhou, and
  Bao]{Shuai22siggraph}
Qing Shuai, Chen Geng, Qi Fang, Sida Peng, Wenhao Shen, Xiaowei Zhou, and Hujun
  Bao.
\newblock Novel view synthesis of human interactions from sparse multi-view
  videos.
\newblock In \emph{ACM SIGGRAPH 2022 Conference Proceedings}, pages 1--10,
  2022.

\bibitem[Suh et~al.(2020)Suh, Choi, Ito, Kim, Lee, Seo, Jung, Yeo, Namgung,
  Bong, seok Kim, Park, Kim, Ryu, and Park]{Suh20iscas}
Yunjae Suh, Seungnam Choi, Masamichi Ito, Jeongseok Kim, Youngho Lee, Jongseok
  Seo, Heejae Jung, Dong-Hee Yeo, Seol Namgung, Jongwoo Bong, Jun seok Kim,
  Paul K.~J. Park, Joonseok Kim, Hyunsurk Ryu, and Yongin Park.
\newblock A 1280x960 {D}ynamic {V}ision {S}ensor with a 4.95-$\mu$m pixel pitch
  and motion artifact minimization.
\newblock In \emph{{IEEE} Int. Symp. Circuits Syst. (ISCAS)}, pages 1--5, 2020.

\bibitem[Sun et~al.(2017)Sun, Shang, Liang, and Wei]{sun2017compositional}
Xiao Sun, Jiaxiang Shang, Shuang Liang, and Yichen Wei.
\newblock Compositional human pose regression.
\newblock In \emph{Proceedings of the IEEE international conference on computer
  vision}, pages 2602--2611, 2017.

\bibitem[Trumble et~al.(2018)Trumble, Gilbert, Hilton, and
  Collomosse]{trumble2018deep}
Matthew Trumble, Andrew Gilbert, Adrian Hilton, and John Collomosse.
\newblock Deep autoencoder for combined human pose estimation and body model
  upscaling.
\newblock In \emph{Proceedings of the European Conference on Computer Vision
  (ECCV)}, pages 784--800, 2018.

\bibitem[Wang et~al.(2019)Wang, Lin, Jiang, Qian, and Wei]{wang20193d}
Keze Wang, Liang Lin, Chenhan Jiang, Chen Qian, and Pengxu Wei.
\newblock 3d human pose machines with self-supervised learning.
\newblock \emph{IEEE transactions on pattern analysis and machine
  intelligence}, 42\penalty0 (5):\penalty0 1069--1082, 2019.

\bibitem[Wang et~al.(2020)Wang, Idoughi, and Heidrich]{Wang20eccv}
Yuanhao Wang, Ramzi Idoughi, and Wolfgang Heidrich.
\newblock Stereo event-based particle tracking velocimetry for {3D} fluid flow
  reconstruction.
\newblock In \emph{Eur. Conf. Comput. Vis. (ECCV)}, pages 36--53, 2020.

\bibitem[Wang et~al.(2022)Wang, Chaney, and Daniilidis]{Wang2022EvAC3D}
Ziyun Wang, Kenneth Chaney, and Kostas Daniilidis.
\newblock Evac3d: From event-based apparent contours to 3d models via
  continuous visual hulls.
\newblock In \emph{European conference on computer vision}, 2022.

\bibitem[Xu et~al.(2020)Xu, Xu, Golyanik, Habermann, Fang, and
  Theobalt]{eventCap2020CVPR}
Lan Xu, Weipeng Xu, Vladislav Golyanik, Marc Habermann, Lu Fang, and Christian
  Theobalt.
\newblock Eventcap: Monocular 3d capture of high-speed human motions using an
  event camera.
\newblock In \emph{{IEEE} Conference on Computer Vision and Pattern Recognition
  (CVPR)}. {IEEE}, 2020.

\bibitem[Yu et~al.(2018)Yu, Zheng, Guo, Zhao, Dai, Li, Pons-Moll, and
  Liu]{yu2018doublefusion}
Tao Yu, Zerong Zheng, Kaiwen Guo, Jianhui Zhao, Qionghai Dai, Hao Li, Gerard
  Pons-Moll, and Yebin Liu.
\newblock Doublefusion: Real-time capture of human performances with inner body
  shapes from a single depth sensor.
\newblock In \emph{Proceedings of the IEEE conference on computer vision and
  pattern recognition}, pages 7287--7296, 2018.

\bibitem[Zhang et~al.(2021)Zhang, Tian, Zhou, Ouyang, Liu, Wang, and
  Sun]{pymaf2021}
Hongwen Zhang, Yating Tian, Xinchi Zhou, Wanli Ouyang, Yebin Liu, Limin Wang,
  and Zhenan Sun.
\newblock Pymaf: 3d human pose and shape regression with pyramidal mesh
  alignment feedback loop.
\newblock In \emph{Proceedings of the IEEE International Conference on Computer
  Vision}, 2021.

\bibitem[Zhang et~al.(2023)Zhang, Tian, Zhang, Li, An, Sun, and
  Liu]{pymafx2023}
Hongwen Zhang, Yating Tian, Yuxiang Zhang, Mengcheng Li, Liang An, Zhenan Sun,
  and Yebin Liu.
\newblock Pymaf-x: Towards well-aligned full-body model regression from
  monocular images.
\newblock \emph{IEEE Transactions on Pattern Analysis and Machine
  Intelligence}, 2023.

\bibitem[Zou et~al.(2021)Zou, Guo, Zuo, Wang, Wang, Hu, Chen, Gong, and
  Cheng]{zou2021eventhpe}
Shihao Zou, Chuan Guo, Xinxin Zuo, Sen Wang, Pengyu Wang, Xiaoqin Hu, Shoushun
  Chen, Minglun Gong, and Li Cheng.
\newblock Eventhpe: Event-based 3d human pose and shape estimation.
\newblock In \emph{Proceedings of the IEEE/CVF International Conference on
  Computer Vision}, pages 10996--11005, 2021.

\end{thebibliography}

}

\end{document}